\documentclass[10pt,twocolumn,letterpaper]{article}

\usepackage{cvpr}              

\usepackage{graphicx}
\usepackage{amsmath}
\usepackage{amssymb}
\usepackage{booktabs}
\usepackage{makecell}
\usepackage[dvipsnames]{xcolor}
\usepackage{enumitem}
\usepackage{centernot}
\usepackage{amsfonts}
\usepackage{algorithm}
\usepackage{algpseudocode}
\usepackage[normalem]{ulem}
\usepackage{dsfont}
\usepackage{multirow}
\usepackage{cuted}
\usepackage{capt-of}
\usepackage{bbm}
\usepackage{balance}
\usepackage[accsupp]{axessibility}

\usepackage[pagebackref,breaklinks,colorlinks,bookmarks=false,citecolor=blue,linkcolor=blue,urlcolor=blue]{hyperref}
\usepackage{xspace}

\usepackage[pagebackref,breaklinks,colorlinks]{hyperref}
\usepackage{amsmath,amsfonts,bm}

\def\1{\bm{1}}

\def\rvc{{\mathbf{c}}}

\def\rvm{{\mathbf{m}}}

\def\rvs{{\mathbf{s}}}

\def\rvz{{\mathbf{z}}}

\def\ervm{{\textnormal{m}}}

\def\ervs{{\textnormal{s}}}

\def\ervz{{\textnormal{z}}}

\def\rmI{{\mathbf{I}}}

\def\rmV{{\mathbf{V}}}

\def\vzero{{\bm{0}}}

\def\vs{{\bm{s}}}

\DeclareMathAlphabet{\mathsfit}{\encodingdefault}{\sfdefault}{m}{sl}
\SetMathAlphabet{\mathsfit}{bold}{\encodingdefault}{\sfdefault}{bx}{n}

\def\gM{{\mathcal{M}}}

\def\gT{{\mathcal{T}}}
\def\gU{{\mathcal{U}}}

\def\sR{{\mathbb{R}}}

\def\sZ{{\mathbb{Z}}}

\newcommand{\E}{\mathbb{E}}
\newcommand{\Ls}{\mathcal{L}}

\DeclareMathOperator{\allpadded}{ispad}
\DeclareMathOperator{\gumbel}{Gumbel}

\usepackage[capitalize]{cleveref}
\crefname{section}{Section}{Secs.}
\Crefname{section}{Section}{Sections}
\Crefname{table}{Table}{Tables}
\crefname{table}{Tab.}{Tabs.}

\definecolor{mygray}{gray}{0.4}
\newenvironment{mytiny}{\color{mygray}\tiny}{}

\def\cvprPaperID{6087} 
\def\confName{CVPR}
\def\confYear{2023}

\graphicspath{{figures/}}

\begin{document}


\newcommand{\lu}[1]{{\textcolor{red}{[lu: #1]}}}
\newcommand{\lijun}[1]{{\textcolor{orange}{[lijun: #1]}}}

\makeatletter
\DeclareRobustCommand\onedot{\futurelet\@let@token\@onedot}
\def\@onedot{\ifx\@let@token.\else.\null\fi\xspace}

\def\eg{\emph{e.g}\onedot} \def\Eg{\emph{E.g}\onedot}
\def\ie{\emph{i.e}\onedot} \def\Ie{\emph{I.e}\onedot}
\def\cf{\emph{c.f}\onedot} \def\Cf{\emph{C.f}\onedot}
\def\etc{\emph{etc}\onedot} \def\vs{\emph{vs}\onedot}
\def\wrt{w.r.t\onedot} \def\dof{d.o.f\onedot}
\def\etal{\emph{et al}\onedot}
\makeatother

\newcommand{\modelname}{MAGVIT}
\newcommand{\methodname}{COMMIT}
\newcommand{\mask}{\texttt{[MASK]}\xspace}
\newcommand{\keep}{0\xspace}
\newcommand{\intecond}{\texttt{[INCO]}\xspace}
\newcommand{\pad}{\texttt{[PAD]}\xspace}

\title{\modelname: Masked Generative Video Transformer}

\author{Lijun Yu$^{\ddagger \dagger \diamond}$\thanks{Work partially done during a research internship at Google Research.}, Yong Cheng$^{\dagger}$, Kihyuk Sohn$^{\dagger}$, Jos\'e Lezama$^{\dagger}$, Han Zhang$^{\dagger}$, Huiwen Chang$^{\dagger}$,\\Alexander G. Hauptmann$^{\ddagger}$, Ming-Hsuan Yang$^{\dagger}$, Yuan Hao$^{\dagger}$, Irfan Essa$^{\dagger \nmid}$, and Lu Jiang$^{\dagger \diamond}$ \\
$^\ddagger$Carnegie Mellon University, $^\dagger$Google Research, $^\nmid$Georgia Institute of Technology \\
{\small$^\diamond$ {\tt Correspondence to lijun@cmu.edu, lujiang@google.com}}
\vspace{-.5em}\\
}


\maketitle

\begin{strip}\centering
\vspace{-16mm}
\includegraphics[width=\textwidth, trim={0.2cm 0 0 0},clip]{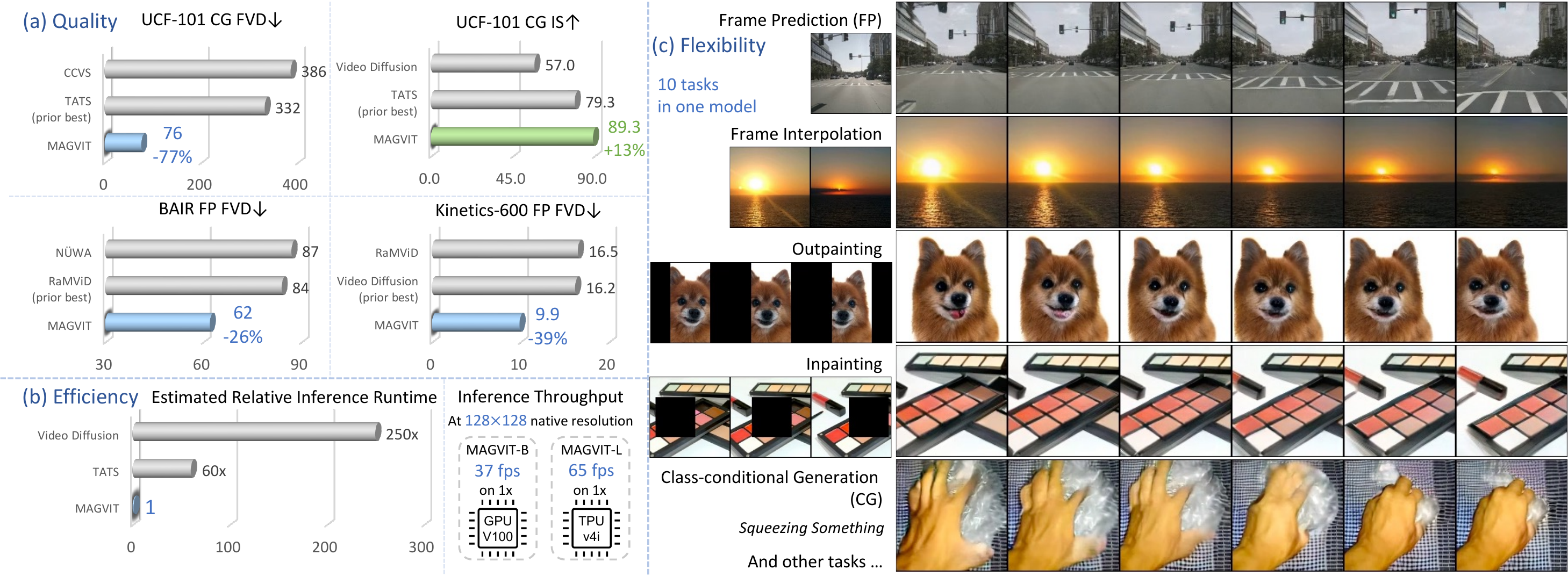}
\vspace{-7mm}
\captionof{figure}{\textbf{Overview of the video generation \emph{quality}, \emph{efficiency}, and \emph{flexibility} of the proposed \modelname{} model.}
%
(a) \modelname{} achieves the state-of-the-art FVD~\cite{unterthiner2018towards} and Inception Score (IS)~\cite{saito2020train} on two video generation tasks and three benchmarks, in comparison with prior best diffusion models (RaMViD~\cite{hoppe2022diffusion}, Video Diffusion~\cite{ho2022video}) and autoregressive models (CCVS~\cite{le2021ccvs}, TATS~\cite{ge2022long}, {N\"UWA}~\cite{wu2021n}).
%
%
(b) It is two orders of magnitude faster than diffusion models and 60$\times$ faster than autoregressive models.  
%
(c) A single \modelname{} model accommodates different generation tasks, ranging from class-conditional generation to dynamic inpainting of a moving object.
}
\label{fig:teaser}
\vspace{-1mm}
\end{strip}

\begin{abstract}
\vspace{-1mm}
We introduce the MAsked Generative VIdeo Transformer, \emph{\modelname{}}, to tackle various video synthesis tasks with a single model.
%
We introduce a 3D tokenizer to quantize a video into spatial-temporal visual tokens and propose an embedding method for masked video token modeling to facilitate multi-task learning.
We conduct extensive experiments to demonstrate the quality, efficiency, and flexibility of \modelname{}.
Our experiments show that (i) \modelname{} performs favorably against state-of-the-art approaches and establishes the best-published FVD on three video generation benchmarks, including the challenging Kinetics-600. 
(ii) \modelname{} outperforms existing methods in inference time by  two orders of magnitude against diffusion models and by 60$\times$ against autoregressive models. 
(iii) A single \modelname{} model supports ten diverse generation tasks and generalizes across videos from different visual domains.
%
The source code and trained models will be released to the public at \url{https://magvit.cs.cmu.edu}.

\vspace{-3mm}
\end{abstract}

\section{Introduction}
\label{sec:intro}

Recent years have witnessed significant advances in image and video content creation based on learning frameworks ranging from
generative adversarial networks (GANs)~\cite{vondrick2016generating,tulyakov2018mocogan,saito2017temporal,clark2019adversarial,luc2020transformation}, 
diffusion models~\cite{rombach2022high,ho2022video,voleti2022masked,hoppe2022diffusion,gu2022vector},
to vision transformers~\cite{weissenborn2019scaling, rakhimov2021latent, nash2022transframer}. 
Inspired by the recent success of generative image transformers such as DALL·E~\cite{ramesh2021zero} and other approaches~\cite{esser2021taming,yu2022scaling,chang2022maskgit,ding2021cogview}, we propose an efficient and effective video generation model by leveraging masked token modeling and multi-task learning.

We introduce the MAsked Generative VIdeo Transformer (\emph{\modelname{}}) for multi-task video generation. 
Specifically, we build and train a single \modelname{} model to perform a variety of diverse video generation tasks and demonstrate the model's efficiency, effectiveness, and flexibility against state-of-the-art approaches.
\cref{fig:teaser}(a) shows the quality metrics of \modelname{} on a few benchmarks with efficiency comparisons in (b), and generated examples under different task setups such as frame prediction/interpolation, out/in-painting, and class conditional generation in (c).

%
\modelname{} models a video as a sequence of visual tokens in the latent space and learns to predict masked tokens with BERT~\cite{devlin2019bert}.
There are two main modules in the proposed framework. 
First, we design a 3D quantization model to tokenize a video, with high fidelity, into a low-dimensional spatial-temporal manifold~\cite{yan2021videogpt, ge2022long}.
Second, we propose an effective \emph{masked token modeling} (MTM) scheme for multi-task video generation.
Unlike conventional MTM in image understanding~\cite{wang2022image} or image/video synthesis~\cite{chang2022maskgit,gupta2022maskvit,han2022show}, we present an embedding method to model a video condition using a multivariate mask and show its efficacy in training. 

We conduct extensive experiments to demonstrate the quality, efficiency, and flexibility of \modelname{} against state-of-the-art approaches. 
Specifically, we show that \modelname{} performs favorably on two video generation tasks
across three benchmark datasets, including UCF-101~\cite{soomro2012ucf101}, BAIR Robot Pushing~\cite{ebert2017self,unterthiner2018towards}, and Kinetics-600~\cite{carreira2018short}.
%
For the class-conditional generation task on UCF-101, \modelname{}
reduces state-of-the-art  FVD~\cite{unterthiner2018towards} 
from $332$~\cite{ge2022long} to $76$ (\mbox{$\downarrow\!77\%$}). 
For the frame prediction task, \modelname{} performs best in terms of FVD on BAIR ($84$  \cite{hoppe2022diffusion} $\!\rightarrow\! 62, \downarrow\!26\%)$ and  Kinetics-600 ($16$ \cite{ho2022video}  $\!\rightarrow\! 9.9, \downarrow\!38\%)$.

Aside from the visual quality, \modelname{}'s video synthesis is highly efficient. 
For instance, \modelname{} generates a 16-frame 128$\times$128 video clip in $12$ steps, which takes $0.25$ seconds on a single TPUv4i~\cite{jouppi2021ten} device. 
On a V100 GPU, a base variant of \modelname{} runs at $37$ frame-per-second (fps) at 128$\times$128 resolution.
%
When compared at the same resolution, \modelname{} is two orders of magnitude faster than the video diffusion model~\cite{ho2022video}.
%
In addition, \modelname{} is $60$ times faster than the autoregressive video transformer~\cite{ge2022long} and $4$-$16$ times more efficient than the contemporary non-autoregressive video transformer~\cite{gupta2022maskvit}.
%

We show that \modelname{} is flexible and robust for multiple video generation tasks with a single trained model, including frame interpolation, class-conditional frame prediction, inpainting, and outpainting, etc.
In addition, \modelname{} learns to synthesize videos with complex scenes and motion contents from diverse and distinct visual domains, including 
actions with objects~\cite{goyal2017something}, autonomous driving~\cite{caesar2020nuscenes}, and object-centric videos from multiple views~\cite{ahmadyan2021objectron}.
%
%
%
%
%

The main contributions of this work are:
\begin{itemize}[nosep, leftmargin=*]
    \item To the best of our knowledge, we present the first masked multi-task transformer for efficient video generation and manipulation. We show that a trained model can perform ten different tasks at inference time.
    \item We introduce a spatial-temporal video quantization model design with high reconstruction fidelity.
    \item We propose an effective embedding method with diverse masks for numerous video generation tasks. 
    \item We show that \modelname{} achieves the best-published fidelity on three widely-used benchmarks, including  UCF-101, BAIR Robot Pushing, and Kinetics-600 datasets.
\end{itemize}

\begin{figure*}[!t]
\centering
\includegraphics[width=\linewidth,trim={0.2cm 0 0 0},clip]{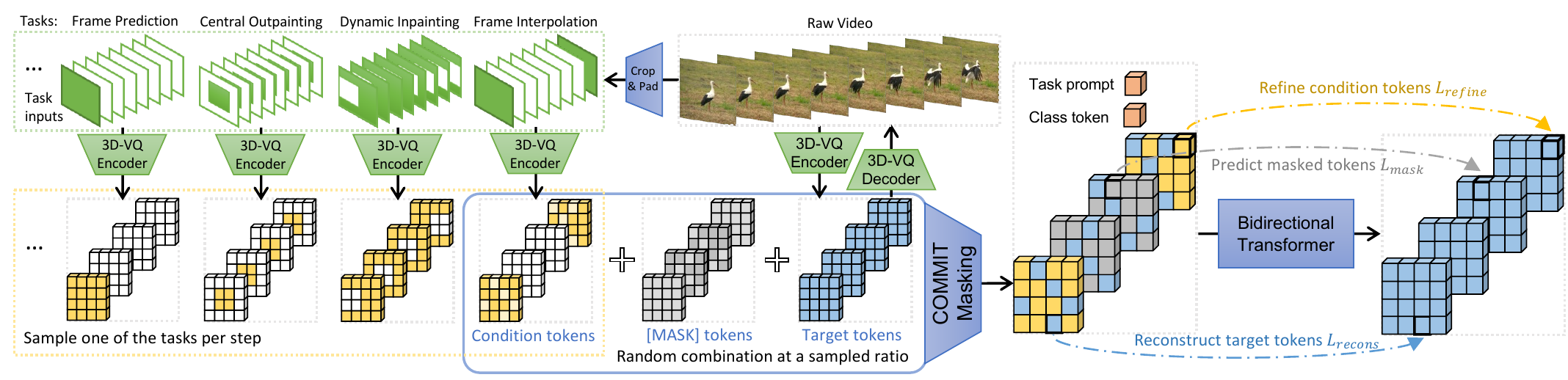}
\vspace{-5mm}
\caption{\textbf{\modelname{} pipeline overview.} The 3D-VQ encoder quantizes a video into discrete tokens, while the 3D-VQ decoder maps them back to the pixel space. 
We sample one of the tasks at each training step and build its condition inputs by cropping and padding the raw video, where \textcolor{ForestGreen}{green} denotes valid pixels and white is padding.
We quantize the condition inputs with the 3D-VQ encoder and select the non-padding part as condition tokens. 
The masked token sequence combines \textcolor{YellowOrange}{condition tokens}, \textcolor{gray}{\mask tokens}, and the \textcolor{RoyalBlue}{target tokens}, with a task prompt and a class token as the prefix. 
The bidirectional transformer learns to predict the target tokens through three objectives: \textcolor{YellowOrange}{refining condition tokens}, \textcolor{gray}{predicting masked tokens}, and \textcolor{RoyalBlue}{reconstructing target tokens}.
}
\label{fig:pipeline}
\vspace{-6mm}
\end{figure*}

\section{Preliminaries: Masked Image Synthesis}
\vspace{-1mm}
\label{sec:background}


%
%
The proposed video generation framework is based on a two-stage image synthesis process~\cite{ramesh2021zero,esser2021taming} with non-autoregressive transformers~\cite{chang2022maskgit,lezama2022improved}.
In the first stage, an image is quantized and flattened into a sequence of discrete tokens by a Vector-Quantized (VQ) auto-encoder~\cite{van2017neural,esser2021taming,yu2021vector}. 
In the second stage, masked token modeling (MTM) is used to train a transformer model~\cite{gu2022vector,chang2022maskgit} on the tokens.
Let $\rmI \in \sR^{H \times W \times 3}$ be an image and $\rvz \in \sZ^N$ denote the corresponding token sequence of length $N$.

We take MaskGIT~\cite{chang2022maskgit} as an example.
In the second stage, it applies a binary mask 
$\ervm_i \in \{x\!\rightarrow\!x, x\!\rightarrow\!\mask\}$
to each token to build a corrupted sequence 
$\overline{\rvz}=\rvm(\rvz)$.
%
Condition inputs, such as class labels, are incorporated as the prefix tokens $\rvc$.
A BERT~\cite{devlin2019bert}  parameterized by $\theta$ is learned to predict the masked tokens in the input sequence $[\rvc, \overline{\rvz}]$, where $[\cdot,\cdot]$ concatenates the sequences.
The objective is to minimize the cross-entropy between the predicted and the ground-truth token at each masked position:
\vspace{-2mm}
\begin{equation}
\Ls_{\text{mask}}(\rvz; \theta) = \mathop{\E}\limits_{\rvm \sim p_{\gU}} \Big[\sum_{\overline{\ervz}_i = \mask} - \log p_\theta (\ervz_i \mid [\rvc, \overline{\rvz}]) \Big]
\label{eq:loss_mtm}
\vspace{-2mm}
\end{equation}
During training, MaskGIT randomly samples $\rvm$ 
from a prior distribution $p_{\gU}$ 
where the mask ratio follows a cosine scheduling function $\gamma(\cdot)$~\cite{chang2022maskgit}.
Specifically, it first uniformly samples a per-token mask score $\ervs_i \sim \gU(0, 1)$ to form a sequence denoted as $\rvs$.
Then it samples $r \sim \gU(0, 1)$ and computes a cut-off threshold $s^*$ as the $\lceil \gamma(r)N \rceil$-\textit{th} smallest element in $\rvs$. 
Finally, a mask $\rvm$ is created such that $\ervm_i(x) = \mask$ if $\ervs_i \le s^*$ and $\ervm_i(x) = x$ otherwise.

For inference,
the non-autoregressive decoding method~\cite{ghazvininejad2019mask,gu2021fully,kong2021blt} is used to synthesize an image~\cite{chang2022maskgit,lezama2022improved,zhang2021m6}. 
For example, MaskGIT
generates an image in $K\!\!=\!\!12$ steps~\cite{chang2022maskgit} from a blank canvas with all visual tokens masked out.
At each step, it predicts all tokens in parallel while retaining tokens with the highest prediction scores. 
The remaining tokens are masked and predicted in the next iteration until all tokens are generated. 
Similar to the training stage, the mask ratio is computed by the schedule function $\gamma$, but with a deterministic input as $\gamma(\frac{t}{K})$, where $t$ is the current step.

\vspace{-1mm}
\section{Masked Generative Video Transformer}
\label{sec:method}
%
Our goal is to design a multi-task video generation model with high quality and inference efficiency.
We propose MAsked Generative VIdeo Transformer (\modelname{}), a vision transformer framework that leverages masked token modeling and multi-task learning.
%
\modelname{} generates a video from task-specific condition inputs, such as a frame, a partially-observed video volume, or a class identifier.

The framework consists of two stages.
First, we learn a 3D vector-quantized (VQ) autoencoder to quantize a video into discrete tokens. 
In the second stage, we learn a video transformer by multi-task masked token modeling. 

Fig.~\ref{fig:pipeline} illustrates the training in the second stage.
At each training step, we sample one of the tasks with its prompt token,
obtain a task-specific conditional mask, and optimize the transformer to predict all target tokens given masked inputs.
%
During inference, we adapt the non-autoregressive decoding method to generate tokens conditionally on the task-specific inputs, which will be detailed in \cref{alg:decoding}.
%


\subsection{Spatial-Temporal Tokenization}
\label{sec:mth_tok}
Our video VQ autoencoder is built upon the image VQGAN~\cite{esser2021taming}.
Let $\rmV \in \sR^{T \times H \times W \times 3}$ be a video clip of $T$ frames.
The VQ encoder tokenizes the video as $f_\gT: \rmV \rightarrow \rvz \in \sZ^{N}$, where $\sZ$ is the codebook.
%
%
The decoder $f_\gT^{-1}$ maps the latent tokens back to video pixels.

The VQ autoencoder is a crucial module as it not only sets a quality bound for the generation but also determines the token sequence length, hence affecting generation efficiency. 
Existing methods apply VQ encoders either on each frame independently (2D-VQ)~\cite{le2021ccvs, gupta2022maskvit} or on a supervoxel (3D-VQ)~\cite{yan2021videogpt, ge2022long}.
We propose different designs that facilitate \modelname{} to perform favorably against other VQ models for video (see \cref{tab:token}). 

\vspace{-5mm}
\paragraph{3D architecture.}
%
We design a 3D-VQ network architecture to model the temporal dynamics as follows.
The encoder and decoder of VQGAN consist of cascaded residual blocks~\cite{he2016deep} interleaved by downsampling (average pooling) and upsampling (resizing plus convolution) layers.
We expand all 2D convolutions to 3D convolutions with a temporal axis.
As the overall downsampling rate is usually different between temporal and spatial dimensions,
we use both 3D and 2D downsampling layers, where the 3D ones appear in the shallower layers of the encoder.
The decoder mirrors the encoder with 2D upsampling layers in the first few blocks, followed by 3D ones.
Appendix A.1
illustrates the detailed architecture. 
%
Note that a token is not only correlated to its corresponding supervoxel
but depends on other patches due to the non-local receptive field.

\vspace{-4mm}
\paragraph{Inflation and padding.}
We initialize our 3D-VQ with weights from a 2D-VQ in a matching architecture to transfer learned spatial relationships~\cite{carreira2017quo}, known as 3D inflation.
We use inflation on small datasets such as UCF-101~\cite{soomro2012ucf101}.
We use a central inflation method for the convolution layers, where the corresponding 2D kernel fills in the temporally central slice of a zero-filled 3D kernel.
The parameters of the other layers are directly copied.
%
%
%
To improve token consistency for the same content at different locations~\cite{ge2022long}, we replace the \texttt{same} (zero) padding in the convolution layers with \texttt{reflect} padding, which pads with non-zero values. 
%

\vspace{-4mm}
\paragraph{Training.}
We apply the image perceptual loss~\cite{esser2021taming} on each frame. 
The LeCam regularization~\cite{tseng2021regularizing} is added to the GAN loss to improve the training stability. 
We adopt the discriminator architecture from StyleGAN~\cite{karras2019style} and inflate it to 3D.
%
%
With these components, unlike VQGAN, our model is trained stably with GAN loss from the beginning.


\subsection{Multi-Task Masked Token Modeling}
\label{sec:mth_cmm}


In \modelname{}, we adopt various masking schemes to facilitate learning for video generation tasks with different conditions.
The conditions can be a spatial region for inpainting/outpainting or a few frames for frame prediction/interpolation. 
%
We refer to these partially-observed video conditions as \emph{interior conditions}.

We argue that it is suboptimal to directly unmask the tokens corresponding to the region of the interior condition~\cite{chang2022maskgit}. 
%
%
As discussed in \cref{sec:mth_tok}, the non-local receptive field of the tokenizer can leak the ground-truth information into the unmasked tokens,
leading to problematic non-causal masking and poor generalization.

We propose a method, COnditional Masked Modeling by Interior Tokens (or \emph{\methodname{}} for short), to embed interior conditions inside the corrupted visual tokens.
%



\vspace{-4mm}
\paragraph{Training.}
Each training example includes a video $\rmV$ and the optional class annotation $\rvc$.
The target visual tokens come from the 3D-VQ as \textcolor{RoyalBlue}{$\rvz = f_\gT(\rmV)$}.
At each step, we sample a task prompt $\rho$, obtain the task-specific interior condition pixels, pad it into $\tilde{\rmV}$ with the same shape as $\rmV$, and get the condition tokens \textcolor{YellowOrange}{$\tilde{\rvz} = f_\gT(\tilde{\rmV})$}.
%
Appendix B.1
lists the padding functions for each task.

At a sampled mark ratio, we randomly replace target tokens $\textcolor{RoyalBlue}{\ervz_i}$, with either
1) the condition token $\textcolor{YellowOrange}{\tilde{\ervz}_i}$, if the corresponding supervoxel of $\textcolor{RoyalBlue}{\ervz_i}$ contains condition pixels; or
2) the special \mask token, otherwise. 
Formally, we compute the \emph{multivariate} conditional mask 
$\rvm(\cdot \mid \tilde{\rvz})$
as 
\vspace{-2mm}
\begin{equation}
\ervm(\ervz_i \mid \tilde{\ervz}_i) = 
\begin{cases}
\textcolor{YellowOrange}{\tilde{\ervz}_{i}} & \text{if } \ervs_{i} \le s^* \land \neg \allpadded(\tilde{\ervz}_{i}) \\
\textcolor{gray}{\mask} & \text{if } \ervs_{i} \le s^* \land \allpadded(\tilde{\ervz}_{i}) \\
\textcolor{RoyalBlue}{\ervz_i} &\text{if } \ervs_{i} > s^*
\end{cases}
\label{eq:icmask}
\vspace{-2mm}
\end{equation}
where $\ervs_{i}$ and $s^*$ are the per-token mask score and the cut-off score introduced in \cref{sec:background}.
$\allpadded(\tilde{\ervz}_{i})$ returns whether the corresponding supervoxel of $\tilde{\ervz}_{i}$ in $\tilde{\rmV}$ only contains padding.

\cref{eq:icmask} indicates that 
\methodname{} embeds interior conditions as corrupted visual tokens into the multivariate mask 
$\rvm$, which follows a new distribution $p_\gM$ instead of the prior $p_{\gU}$ for binary masks.
With the corrupted token sequence $\overline{\rvz}=\rvm(\rvz \mid \tilde{\rvz})$ as input,
%
%
the \emph{multi-task} training objective is
\vspace{-2mm}
\begin{equation}
\Ls(\rmV; \theta) \!=\!\! \mathop{\E}\limits_{\rho, \widehat{\rmV}} \mathop{\E}\limits_{\rvm \sim p_\gM} \Big[\sum_{i} -\log p_\theta (\ervz_i \mid [\rho, \rvc, \overline{\rvz}]) \Big]
\label{eq:mt_loss}
\vspace{-2mm}
\end{equation}
%
We can decompose the loss in \cref{eq:mt_loss} into three parts according to \cref{eq:icmask}: $\Ls_\text{refine}$ refines the task-specific condition tokens, $\Ls_\text{mask}$ predicts masked tokens , and $\Ls_\text{recons}$ reconstructs target tokens.
Let $\overline{\rvc} = [\rho, \rvc, \overline{\rvz}]$ for simplicity,
\vspace{-4mm}
\begin{equation}
\begin{aligned}
\sum_{i=1}^N - \log p_\theta (\ervz_i \mid [\rho, \rvc,\overline{\rvz}]) &= \!\!\!\!\underbrace{\sum_{\overline{\ervz}_i = \tilde{\ervz}_i}\!\! - \log p_\theta (\ervz_i\mid\overline{\rvc})}_{\text{\textcolor{YellowOrange}{Refine condition tokens $\Ls_\text{refine}$}}} \\
+\!\!\!\underbrace{\sum_{\overline{\ervz}_i = \mask}\!\!\!\!\!\!\! - \log p_\theta (\ervz_i\mid\overline{\rvc})}_{\text{\textcolor{gray}{Predict masked tokens $\Ls_\text{mask}$}}} 
\ \ &+\underbrace{\sum_{\overline{\ervz}_i = \ervz_i}\! - \log p_\theta (\ervz_i \mid\overline{\rvc})}_{\text{\textcolor{RoyalBlue}{Reconstruct target tokens $\Ls_\text{recons}$}}} 
\end{aligned}
\label{eq:mt_loss_decomposed}
\vspace{-2mm}
\end{equation}
%
%
While $\Ls_\text{mask}$ is the same as the MTM loss in \cref{eq:loss_mtm} and $\Ls_\text{recons}$ sometimes is used as a regularizer (\eg, in NLP tasks), $\Ls_\text{refine}$ is a new component introduced by \methodname{}. 

The  \methodname{} method facilitates multi-task video generation in three aspects. 
First, it provides a correct causal masking for all interior conditions.
Second, it produces a fixed-length sequence for different conditions of arbitrary regional volume, improving training and memory efficiency since no padding tokens are needed.
%
Third, it achieves state-of-the-art multi-task video generation results (see \cref{tab:cond}).

\vspace{-4mm}
\paragraph{Video generation tasks.} 
We consider \emph{ten} tasks for multi-task video generation where each task has a different interior condition and mask:
Frame Prediction (FP), 
Frame Interpolation (FI),
Central Outpainting (OPC),
Vertical Outpainting (OPV),
Horizontal Outpainting (OPH),
Dynamic Outpainting (OPD),
Central Inpainting (IPC),
and Dynamic Inpainting (IPD),
Class-conditional Generation (CG), 
Class-conditional Frame Prediction (CFP). 
We provide the detailed definitions in Appendix B.1.

\begin{algorithm}[tp]
\caption{Non-autoregressive Decoding by \methodname{}}
\label{alg:decoding}
\textbf{Input}: prefix $\rho$ and $\rvc$, condition $\tilde{\rvz}$, steps $K$, temperature $T$ \\
\textbf{Output}: predicted visual tokens $\hat{\rvz}$
\begin{algorithmic}[1]
{\small
\State $\rvs = \vzero$, $s^*=1$, $\hat{\rvz} = \vzero^N$
\For {$t \gets 0, 1, \cdots, K - 1$}
\State $\overline{\rvz} \gets \rvm(\hat{\rvz} \mid \tilde{\rvz}; \rvs, s^*)$
\State $\hat{\ervz}_{i} \sim p_{\theta}(\ervz_i \mid [\rho, \rvc, \overline{\rvz}])$, $\forall i$ where $\ervs_i \le s^*$
\State $\ervs_i \gets p_\theta(\hat{\ervz}_i \mid [\rho, \rvc, \overline{\rvz}])$, $\forall i$ where $\ervs_i \le s^*$
\State $\ervs_i \gets \ervs_i + T (1 - \frac{t+1}{K})\gumbel(0, 1)$, $\forall i$ where $\ervs_i < 1$
\State $s^* \gets$ The $\lceil \gamma(\frac{t+1}{K})N \rceil$-\textit{th} smallest value of $\rvs$
\State $\ervs_i \gets 1$, $\forall i$ where $\ervs_i > s^*$
\EndFor \\
\Return $\hat{\rvz} = [\hat{\ervz}_1, \hat{\ervz}_2, \cdots, \hat{\ervz}_N]$
}
\end{algorithmic}
\end{algorithm}

\begin{figure}[tp]
\centering
\vspace{-1mm}
\includegraphics[width=\linewidth,trim={0.25cm 0.2cm 0.1cm 0.2cm},clip]{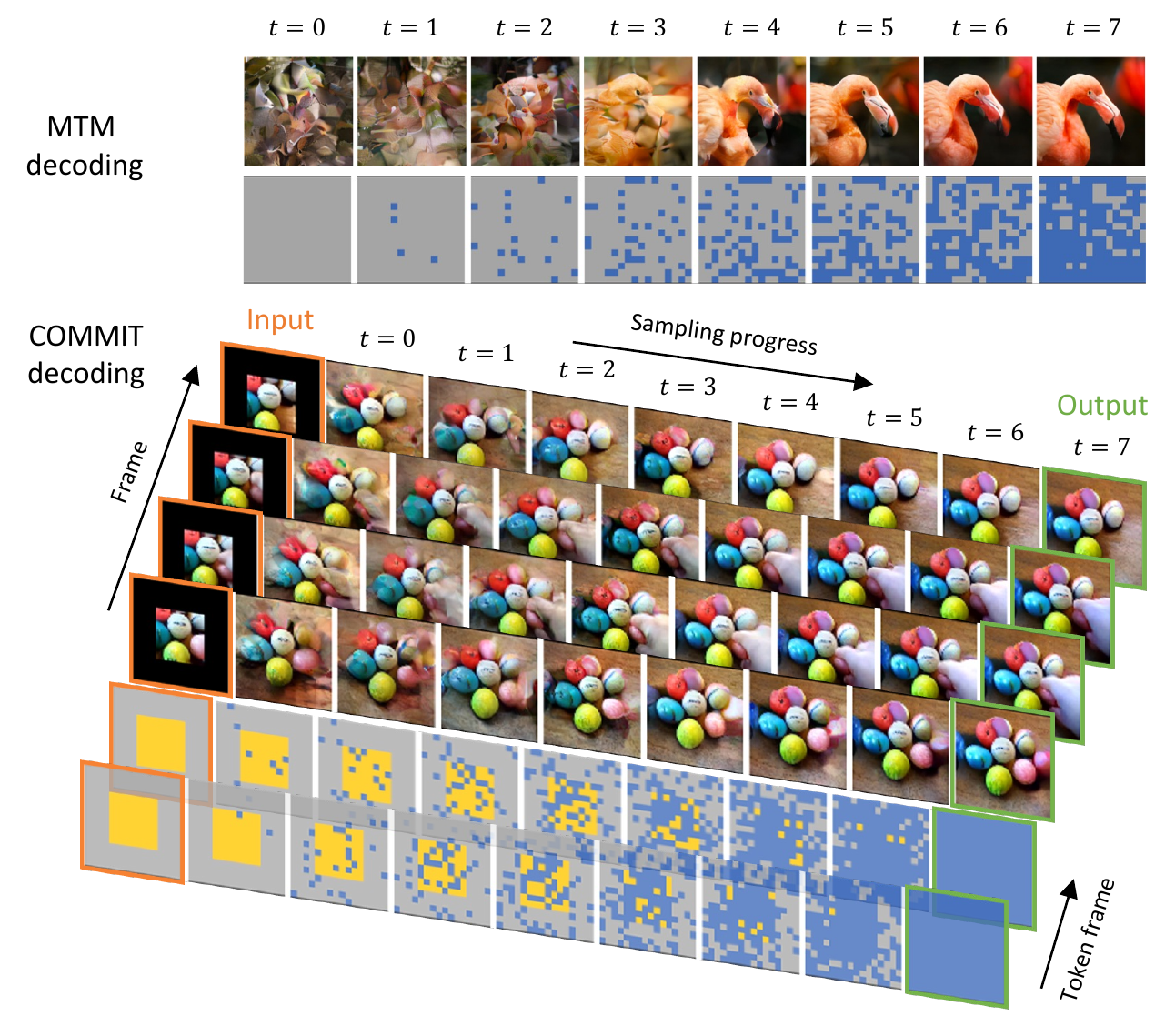}
\caption{\textbf{Comparison between MTM decoding for image~\cite{chang2022maskgit} and \methodname{} decoding for video}. We show the output tokens and image/video at each decoding step $t$, with a central outpainting example for \methodname{}.
Unlike the MTM denoising decoding from all \textcolor{gray}{\mask}, \methodname{} performs a conditional generation process toward the \textcolor{RoyalBlue}{output tokens} while gradually replacing the \textcolor{YellowOrange}{interior condition tokens}.
Videos and tokens are temporally down-sampled and stacked for visualization.
}
\label{fig:sample}
\vspace{-6mm}
\end{figure}
%
\vspace{-4mm}
\paragraph{Inference.}
We use a non-autoregressive decoding method to generate video tokens from input conditions in $K$ steps (\eg, $12$).
Each decoding step follows the \methodname{} masking in \cref{eq:icmask} with a gradually reduced mask ratio.
\cref{alg:decoding} outlines the inference procedure.



\cref{fig:sample} compares the non-autoregressive image decoding~\cite{chang2022maskgit} and our video decoding procedure.
Different from the MTM decoding in~\cite{chang2022maskgit} which performs denoising from all \mask, \methodname{} decoding starts from a \emph{multivariate} mask that embeds the \textcolor{YellowOrange}{interior conditions}.
Guided by this mask, \cref{alg:decoding} performs a conditional transition process toward the output tokens by replacing a portion of newly generated tokens at each step.
In the end, all tokens are predicted where the interior condition tokens get refined. 



\section{Experimental Results}
\label{sec:experiments}

We conduct extensive experiments to demonstrate the video generation quality (\cref{sec:exp_quality}), efficiency (\cref{sec:exp_efficiency}), and flexibility for multi-task generation (\cref{sec:exp_flexibility}).
We show a few generation results here, and refer to the
\iftoggle{cvprfinal}{%
web page\footnote{\url{https://magvit.cs.cmu.edu}}
}{%
anonymous page\footnote{\url{https://magvit-ann.github.io}}
}%
for more examples. 

\subsection{Experimental Setups}

\begin{table}[tp]
\centering
\scriptsize
\begin{tabular}{@{}lcccc@{}}
\toprule
Method        & Extra Video  & Class      & FVD$\downarrow$          & IS$\uparrow$                \\ \midrule
RaMViD\cite{hoppe2022diffusion}    & &    &   -  &  21.71\mytiny{$\pm$0.21}  \\
StyleGAN-V$^{*}$\cite{skorokhodov2022stylegan} & & &-& 23.94\mytiny{$\pm$0.73} \\
DIGAN\cite{yu2022generating} & &            & 577\mytiny{$\pm$21}  & 32.70\mytiny{$\pm$0.35} \\
DVD-GAN\cite{clark2019adversarial} & & \checkmark & -                        & 32.97\mytiny{$\pm$1.70}  \\
Video Diffusion$^*$\cite{ho2022video} &  &            & -                        & 57.00\mytiny{$\pm$0.62}    \\
TATS\cite{ge2022long}  & &            & 420\mytiny{$\pm$18}  & 57.63\mytiny{$\pm$0.24} \\
CCVS+StyleGAN\cite{le2021ccvs} & &            & 386\mytiny{$\pm$15}  & 24.47\mytiny{$\pm$0.13} \\
Make-A-Video$^*$\cite{singer2022make} & & \checkmark & 367 & 33.00 \\
TATS\cite{ge2022long} & & \checkmark & 332\mytiny{$\pm$18}  & 79.28\mytiny{$\pm$0.38} \\ \midrule
\color{mygray}CogVideo$^*$\cite{hong2022cogvideo} & \color{mygray}\checkmark  & \color{mygray}\checkmark & \color{mygray}626                      & \color{mygray}50.46                       \\
\color{mygray}Make-A-Video$^*$\cite{singer2022make} & \color{mygray}\checkmark & \color{mygray}\checkmark & \color{mygray}81 & \color{mygray}82.55 \\ \midrule
\emph{\modelname}-B-CG (ours) &  & \checkmark & \underline{159\mytiny{$\pm$2}} & \underline{83.55\mytiny{$\pm$0.14}}     \\
\emph{\modelname}-L-CG (ours) & & \checkmark & \textbf{76\mytiny{$\pm$2}}     & \textbf{89.27\mytiny{$\pm$0.15}}     \\ \bottomrule
\end{tabular}
\vspace{-1mm}
\caption{\textbf{Generation performance on the UCF-101 dataset.} 
Methods in \textcolor{mygray}{gray} are pretrained on additional large video data.
Methods with $\checkmark$ in the Class column are class-conditional, while the others are unconditional.
Methods marked with $^*$ use custom resolutions, while the others are at 128$\times$128. 
See Appendix C for more comparisons with earlier works.}
\label{tab:ucf}
\vspace{-4mm}
\end{table}

\paragraph{Datasets.}
We evaluate the single-task video generation performance of \modelname{} on three standard benchmarks, \ie, class-conditional generation on UCF-101~\cite{soomro2012ucf101} and frame prediction on BAIR Robot Pushing~\cite{ebert2017self, unterthiner2018towards} (1-frame condition) and Kinetics-600~\cite{carreira2018short} (5-frame condition).
For multi-task video generation, we quantitatively evaluate \modelname{} on BAIR and SSv2~\cite{goyal2017something} on 8-10 tasks.
Furthermore, to evaluate model generalizability, we train models with the same learning recipe on three additional video datasets: nuScenes~\cite{caesar2020nuscenes}, Objectron~\cite{ahmadyan2021objectron}, and 12M Web videos. We show their generated videos in the main paper and quantitative performance in Appendix C.


\vspace{-4mm}
\paragraph{Evaluation metrics.}
We use FVD~\cite{unterthiner2018towards} as our primary evaluation metric. 
Similar to~\cite{ge2022long, ho2022video}, FVD features are extracted with an I3D model trained on Kinetics-400~\cite{carreira2017quo}. 
We also report the Inception Score (IS)~\cite{saito2020train} 
calculated with a C3D~\cite{tran2015learning} model on UCF-101, and PSNR, SSIM~\cite{wang2004image}, and LPIPS~\cite{zhang2018unreasonable} on BAIR. 
We report the mean and standard deviation for each metric calculated over four runs. 



\vspace{-4mm}
\paragraph{Implementation details.}
We train \modelname{} to generate 16-frame videos at 128$\times$128 resolution, except for BAIR at 64$\times$64.
The proposed 3D-VQ model quantizes a video into 4$\times$16$\times$16 visual tokens,
where the visual codebook size is 1024.
%
%
%
%
%
We use the BERT transformer~\cite{devlin2019bert} to model the token sequence, which 
includes 1 task prompt, 1 class token, and 1024 visual tokens.
Two variants of \modelname{}, \ie, base (B) with 128M parameters and large (L) with 464M, are evaluated. 
We train both stages with the Adam optimizer~\cite{kingma2014adam} in JAX/Flax~\cite{jax2018github,flax2020github} on TPUs.
Appendix B.2
details training configurations.



\subsection{Single-Task Video Generation}
\label{sec:exp_quality}
\paragraph{Class-conditional generation.}

The model is given a class identifier in this task to generate the full video.
\cref{tab:ucf} shows that \modelname{} surpasses the previous best-published FVD and IS scores. Notably, it outperforms Make-A-Video~\cite{singer2022make} which is pretrained on additional 10M videos with a text-image prior. In contrast, \modelname{} is just trained on the 9.5K training videos of UCF-101.

\begin{figure*}[tp]
\centering
\begin{subfigure}[t]{0.33\linewidth}
\includegraphics[width=\linewidth]{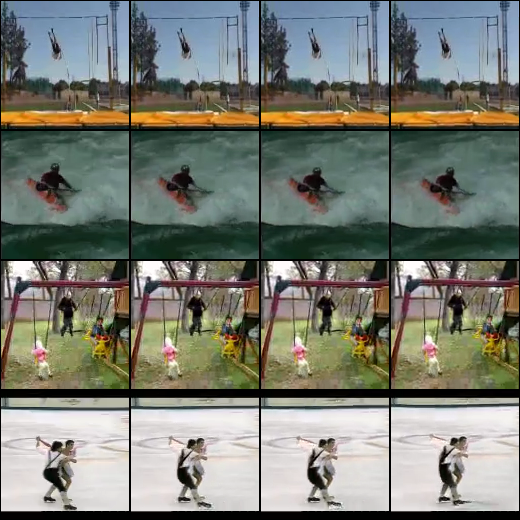}
\caption{CCVS+StyleGAN~\cite{le2021ccvs}
}
\end{subfigure}
\hfill
\begin{subfigure}[t]{0.33\linewidth}
\includegraphics[width=\linewidth]{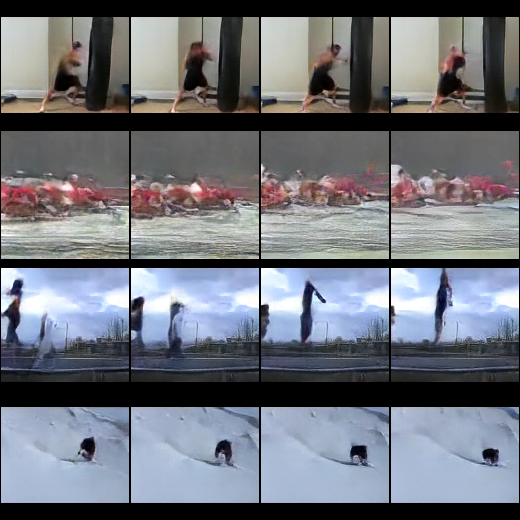}
\caption{TATS~\cite{ge2022long}
}
\end{subfigure}
\hfill
\begin{subfigure}[t]{0.33\linewidth}
\includegraphics[width=\linewidth]{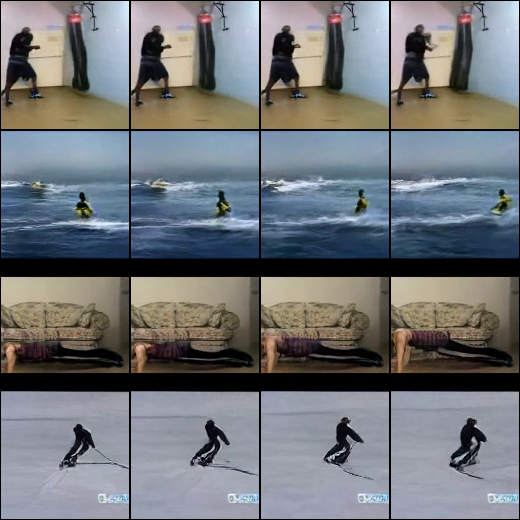}
\caption{\modelname{}-L-CG (ours)
}
\end{subfigure}
\hfill
\vspace{-2mm}
\caption{\textbf{Comparison of class-conditional generation samples on UCF-101.} 16-frame videos are generated at 128$\times$128 resolution 25 fps and shown at 6.25 fps. Samples for \cite{le2021ccvs,ge2022long} are obtained from their official release\protect\footnotemark. More comparisons are provided in Appendix D.}
\label{fig:ucf}
\vspace{-4mm}
\end{figure*}

\cref{fig:ucf} compares the generated videos to baseline models.
We can see that 
CCVS+StyleGAN~\cite{le2021ccvs} gets a decent single-frame quality,
but yields little or no motion.
TATS~\cite{ge2022long} generates some motion but with artifacts.
In contrast, our model produces higher-quality frames with substantial motion.

\vspace{-4mm}
\paragraph{Frame prediction.}
\begin{table}[tp]
\centering
\scriptsize
\begin{tabular}{@{}lcc@{}}
\toprule
Method                   & K600 FVD$\downarrow$   & BAIR FVD$\downarrow$ \\ \midrule
CogVideo\cite{hong2022cogvideo}  & 109.2                      & -                      \\
CCVS\cite{le2021ccvs}    & 55.0\mytiny{$\pm$1.0}  & 99\mytiny{$\pm$2}  \\
Phenaki\cite{villegas2022phenaki} & 36.4\mytiny{$\pm$0.2} & 97 \\
TrIVD-GAN-FP\cite{luc2020transformation} & 25.7\mytiny{$\pm$0.7}  & 103                    \\
Transframer\cite{nash2022transframer}  & 25.4                       & 100                    \\
MaskViT\cite{gupta2022maskvit} & -                          & 94                     \\
FitVid\cite{babaeizadeh2021fitvid} & -                          & 94                     \\
MCVD\cite{voleti2022masked} & - & 90 \\
N\"UWA\cite{wu2021n}    & -                          & 87                     \\
RaMViD\cite{hoppe2022diffusion}        &  16.5     &   84 \\
Video Diffusion\cite{ho2022video}  & \underline{16.2\mytiny{$\pm$0.3}}  & -                     \\ \midrule
\emph{\modelname}-B-FP (ours)          & 24.5\mytiny{$\pm$0.9}     & \underline{76{\mytiny$\pm$0.1}} (48{\mytiny$\pm$0.1})   \\
\emph{\modelname}-L-FP (ours)         & \textbf{9.9\mytiny{$\pm$0.3}}     & \textbf{62{\mytiny$\pm$0.1}}  (31{\mytiny$\pm$0.2})   \\ \bottomrule
\end{tabular}
\vspace{-1mm}
\caption{\textbf{Frame prediction performance on the BAIR and Kinetics-600 datasets.} 
- marks that the value is unavailable in their paper or incomparable to others.
The FVD in parentheses uses a debiased evaluation protocol on BAIR detailed in Appendix B.3.
See Appendix C for more comparisons with earlier works.
}
\label{tab:bairk600}
\end{table}

\begin{table}[tp]
\centering

\scriptsize
\begin{tabular}{@{}lcccccc@{}}
\toprule
Method     & FVD$\downarrow$  & PSNR$\uparrow$ & SSIM$\uparrow$  & LPIPS$\downarrow$ \\ \midrule
CCVS\cite{le2021ccvs} & 99  & - & 0.729 & -  \\
MCVD\cite{voleti2022masked} & 90 & 16.9 & 0.780 & -       \\ \midrule
\emph{\modelname}-L-FP (ours) & \textbf{62} & \textbf{19.3}  &   \textbf{0.787}   & 0.123  \\ \bottomrule
\end{tabular}
\vspace{-1mm}
\caption{\textbf{Image quality metrics on BAIR frame prediction.} 
}
\label{tab:im_quality}
\vspace{-5mm}
\end{table}

The model is given a single or a few frames to generate future frames.
%
In \cref{tab:bairk600}, we compare \modelname{} against highly-competitive baselines. 
\modelname{} surpasses the previous state-of-the-art FVD on BAIR by a large margin ($84 \xrightarrow[]{} 62$). Inspired by~\cite{unterthiner2018towards}, a ``debiased'' FVD is also reported in the parentheses to overcome the small validation set. See more discussion in
Appendix B.3.
In \cref{tab:im_quality}, it demonstrates better image quality.

On the large dataset of Kinetics-600, it establishes a new state-of-the-art result, improving the previous best FVD in~\cite{ho2022video} from $16.2$ to $9.9$ by a relative $39\%$ improvement. The above results verify \modelname{}'s compelling generation quality, including on the large Kinetics dataset.

\subsection{Inference-Time Generation Efficiency}
\label{sec:exp_efficiency}
\begin{figure}[tp]
\centering
\includegraphics[width=\linewidth,trim={0cm 0.25cm 0cm 0.2cm},clip]{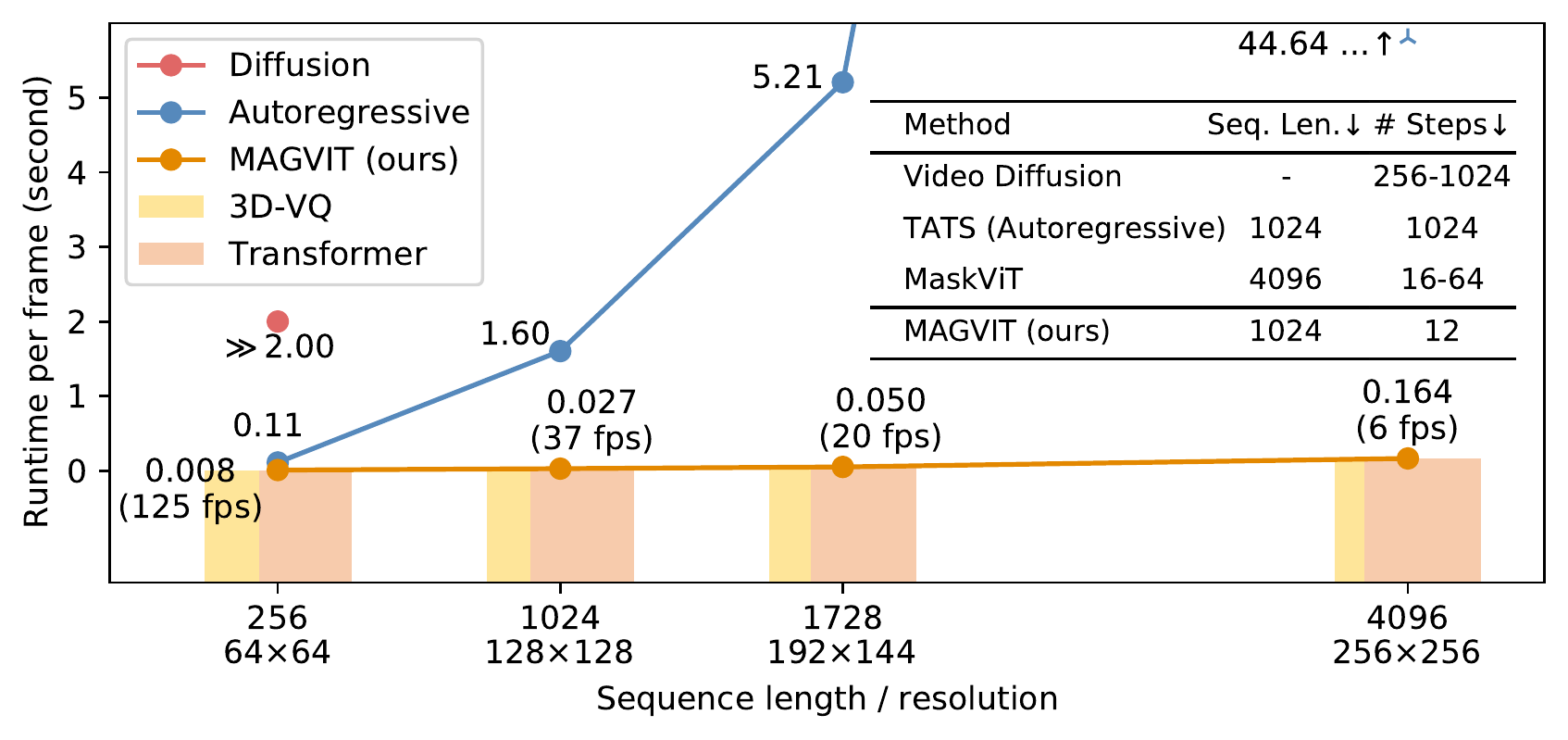}
\vspace{-5mm}
\caption{\textbf{Inference-time generation efficiency comparison.} 
The average runtime for generating one frame is measured at different resolutions. 
The colored bars show the time breakdown between the 3D-VQ and the transformer.
The embedded table compares the critical factors of inference efficiency for different methods at 16-frame 128$\times$128, except for Video Diffusion~\cite{ho2022video} at 64$\times$64.
}
\label{fig:efficiency}
\vspace{-5mm}
\end{figure}

Video generation efficiency is an important metric in many applications. We conduct experiments to validate that \modelname{} offers top speed in video generation.
\cref{fig:efficiency} shows the processing time for each frame on a single V100 GPU at different resolutions.
We compare \modelname-B with an autoregressive transformer of the same size and a diffusion-based model~\cite{ho2022video}.
At 128$\times$128 resolution, \modelname-B runs at $37$ frames-per-second (fps).
When running on a single TPUv4i~\cite{jouppi2021ten},
\modelname{}-B runs at $190$ fps and \modelname{}-L runs at $65$ fps.

\cref{fig:efficiency} compares the sequence lengths and inference steps of these models.
Diffusion models~\cite{ho2022video} typically require 256-1000 diffusion steps with a 3D U-Net~\cite{cciccek20163d}.
Autoregressive models, such as TATS~\cite{ge2022long}, decode visual tokens sequentially, which runs $60$ times slower than \modelname{} at 128$\times$128. 
Compared to the recent non-autoregressive model MaskViT~\cite{gupta2022maskvit}, \modelname{} is $4$ to $16$ times faster due to more efficient decoding on shorter sequences.
\footnotetext{\url{https://songweige.github.io/projects/tats/}}

\begin{table*}[tp]
\centering
\scriptsize
\begin{tabular}{@{}lc|ccccccccc|cccc@{}}
\toprule
Method &Task &\textbf{BAIR-MT8}$\downarrow$ & FP & FI & OPC & OPV & OPH & OPD & IPC & IPD & \textbf{SSV2-MT10}$\downarrow$ & CG & CFP \\ \midrule
\modelname-B-UNC &Single& 150.6 & {\color{mygray}74.0} & {\color{mygray}71.4} & {\color{mygray}119.0} & {\color{mygray}46.7} & {\color{mygray}55.9} & {\color{mygray}389.3} & {\color{mygray}145.0} & {\color{mygray}303.2} & 258.8 & {\color{mygray}107.7} & {\color{mygray}279.0}  \\
\modelname-B-FP  & Single &201.1 & 47.7 & {\color{mygray}56.2}& {\color{mygray}247.1} & {\color{mygray}118.5} & {\color{mygray}142.7}  & {\color{mygray}366.3} & {\color{mygray}357.3} & {\color{mygray}272.7} & 402.9 & {\color{mygray}1780.0} & 59.3 \\ \midrule
\modelname-B-\emph{MT} & Multi& 32.8 & 47.2 & 36.0 & 28.1 & 29.0 & 27.8 & 32.1 & 31.1 & 31.0 & 43.4 & 94.7 & 59.3     \\
\modelname-L-\emph{MT} & Multi& \textbf{22.8} & 31.4 & 26.4 & 21.3 & 21.2 & 19.5 & 20.9 & 21.3 & 20.3 & \textbf{27.3} & 79.1 & 28.5  \\\bottomrule
\end{tabular}
\vspace{-1mm}
\caption{\textbf{Multi-task generation performance on BAIR and SSV2 evaluated by FVD.} \textcolor{mygray}{Gray} values denote unseen tasks during training. We list per-task FVD for all eight tasks on BAIR and the two extra tasks on SSV2 here, and leave the details for SSV2 in Appendix C.}
\label{tab:mt_bair}
\end{table*}


\begin{figure*}[tp]
\centering
\includegraphics[width=\linewidth,trim={0.8cm 0 0 0},clip]{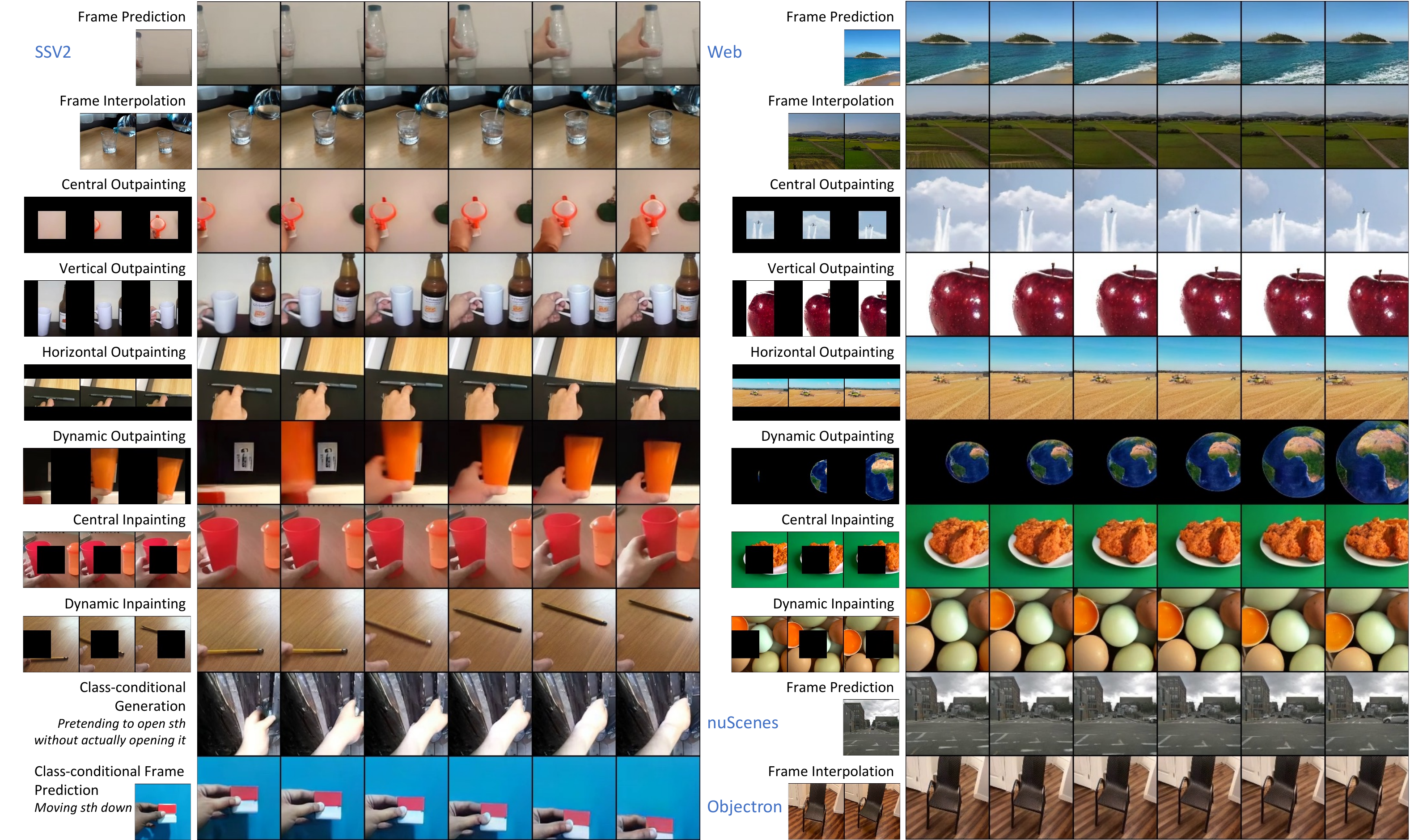}
\vspace{-3mm}
\caption{\textbf{Multi-task generation samples on four datasets: SSv2~\cite{goyal2017something}, nuScenes~\cite{caesar2020nuscenes}, Objectron~\cite{ahmadyan2021objectron}, and Web videos.} The left column is from a single ten-task model on SSv2, while the top eight rows on the right are from a single eight-task model on Web data.}
\label{fig:mt_result}
\vspace{-3mm}
\end{figure*}

\subsection{Multi-task Video Generation}
\label{sec:exp_flexibility}
To demonstrate the flexibility in multi-task video synthesis, we train a single \modelname{} model to perform eight tasks on BAIR or ten tasks on SSv2. 
We do not intend to compare with dedicated models trained on these tasks but to demonstrate a generic model for video synthesis.

\vspace{-4mm}
\paragraph{Eight tasks on BAIR.}
We perform a multi-task evaluation on BAIR with eight self-supervised tasks.
\cref{tab:mt_bair} lists the ``debiased'' FVD for each task, where the third column computes the average. 
We compare the multi-task models (MT) with two single-task baselines trained on unconditional generation (UNC) and frame prediction (FP).

As shown in \cref{tab:mt_bair}, the multi-task models achieve better fidelity across all tasks. Single-task models perform considerably worse on the tasks unseen in training (\textcolor{mygray}{gray} values in \cref{tab:mt_bair}), especially on the tasks that differ more from the training task. 
Compared to the single-task models in their training task, MT performs better with a small gain on FP with the same model size. 

\vspace{-4mm}
\paragraph{Ten tasks on SSv2.}
We evaluate on the large-scale SSv2 dataset, where \modelname{} needs to synthesize 174 basic actions with everyday objects. We evaluate a total of ten tasks, with two of them using class labels (CG and CFP), as shown on the right side of \cref{tab:mt_bair}. 
%
We observe a pattern consistent with BAIR: multi-task models achieve better average FVD across all tasks.
The above results substantiate model generalization trained with the proposed multi-task objective.


\begin{table}[tp]
\centering
\scriptsize
\begin{tabular}{@{}l@{\hspace{5pt}}l@{\hspace{5pt}}c@{\hspace{6pt}}ccc@{}}
\toprule
\multicolumn{2}{l}{Method} & \makecell{Seq. Length} & \makecell{FP FVD$\downarrow$} & \makecell{MT8 FVD$\downarrow$} \\ \midrule
\multicolumn{2}{l}{Latent masking in MaskGIT~\cite{chang2022maskgit}} & 1024 & 74 & 151 \\
\multicolumn{2}{l}{Prefix condition} & 1024-1792   & 55 & - \\ \midrule
\multicolumn{1}{l}{\multirow{3}{*}{\makecell{\emph{\methodname}\\(ours)}}} & $\Ls_\text{mask}$ & \multirow{3}{*}{1024} & 388 & 143 \\
& $\Ls_\text{mask}+\Ls_\text{recons}$ & & 51 & 53 \\
&$\Ls_\text{mask}+\Ls_\text{recons}+\Ls_\text{refine}$ & & \textbf{48} & \textbf{33} \\ \bottomrule
\end{tabular}
\vspace{-1mm}
\caption{\textbf{Comparison of conditional masked token modeling} on BAIR frame prediction (FP) and eight-task (MT8) benchmarks. - indicates we were not able to train to convergence. 
}
\label{tab:cond}
\vspace{-2mm}
\end{table}

\subsection{Ablation Study}
\label{sec:exp_abl}

\paragraph{Conditional MTM.}
We demonstrate the efficacy of \methodname{} by comparing it with conventional MTM methods, including the
latent masking in MaskGIT for image synthesis~\cite{chang2022maskgit} 
and the commonly-used prefix condition that prepends cropped condition tokens to the input sequence.

\cref{tab:cond} compares these methods on the BAIR dataset where the same 3D-VQ tokenizer is used in all approaches. 
As discussed in \cref{sec:mth_cmm}, latent masking in~\cite{chang2022maskgit}, which directly unmasks tokens of the condition region at inference time, leads to poor generalization, especially for the multi-task setup.
Prefix condition produces a long sequence of variable length, making it less tractable for multi-task learning.
In contrast, \methodname{} yields a fixed-length sequence and better generalizability for both single- and multi-task setups.

\vspace{-4mm}
\paragraph{Training losses.} The bottom section of \cref{tab:cond} shows the contribution of the training loss components in Eq.~\eqref{eq:mt_loss_decomposed}. 

\vspace{-4mm}
\paragraph{Decoding methods.}

\begin{table}[tp]
\centering
\scriptsize
\begin{tabular}{@{}l@{\hspace{3pt}}c@{\hspace{5pt}}c@{\hspace{5pt}}c@{\hspace{3pt}}c@{\hspace{3pt}}c@{\hspace{3pt}}c@{\hspace{5pt}}c@{}}
\toprule
Decoding Method & Tokenizer &Type & Param. & Seq. Len.$\downarrow$ & \# Steps$\downarrow$ & \makecell{FVD$\downarrow$} \\ \midrule
\multirow{2}{*}{MaskGIT~\cite{chang2022maskgit}} & 2D-VQ  &NAR& 53M+87M & 4096 & 12 & 222 (177) \\
 & 3D-VQ &NAR& 41M+87M & 1024 & 12 & 122 (74) \\
MaskViT~\cite{gupta2022maskvit} & 2D-VQ  &NAR& 53M+189M & 4096 & 18 & 94$^*$ \\ \midrule 
AR & 3D-VQ &AR&41M+87M & 1024 & 1024 & 91 (56)  \\ \midrule
\emph{\modelname} (ours) & 3D-VQ &NAR&41M+87M & 1024 & 12 & \textbf{76} (\textbf{48})\\ \bottomrule
\end{tabular}
\vspace{-1mm}
\caption{\textbf{Comparison of decoding methods} on BAIR frame prediction benchmark. 
The number of parameters is broken down as VQ + Transformer.
NAR is non-autoregressive and AR is autoregressive. FVD and debiased FVD (in parentheses) are reported.
$^*$ marks the quoted number from their paper.
}
\label{tab:arch}
\vspace{-2mm}
\end{table}

\cref{tab:arch} compares \cref{alg:decoding} with existing autoregressive (AR) and non-autoregressive (NAR) decoding methods. 
We consider two NAR baselines, \ie, MaskGIT~\cite{chang2022maskgit} for image and MaskViT~\cite{gupta2022maskvit} for video synthesis.
%
We use the same 3D-VQ tokenizer for MaskGIT, AR, and \modelname{}. 
As shown, the proposed decoding algorithm produces the best quality with the 3D-VQ and has a $4\times$ shorter sequence than the 2D-VQ.
While the AR transformer obtains a reasonable FVD, it takes 
over $85\times$ more steps at inference time.


\vspace{-4mm}
\paragraph{VQ architecture and training techniques.}

\begin{table}[tp]
\centering
\scriptsize
\begin{tabular}{l@{\hspace{8pt}}c@{\hspace{5pt}}c@{\hspace{8pt}}c@{\hspace{5pt}}c@{\hspace{8pt}}c@{\hspace{5pt}}c}
\toprule
\multirow{2}{*}{Tokenizer} & \multicolumn{2}{l}{From Scratch} & \multicolumn{4}{l}{ImageNet~\cite{deng2009imagenet} Initialization} \\
& FVD$\downarrow$ & IS$\uparrow$ & FVD$\downarrow$ & IS$\uparrow$ & FVD$\downarrow$ & IS$\uparrow$ \\ \midrule
MaskGIT~\cite{chang2022maskgit} 2D-VQ 
 & 240 & 80.9 & 216 & 82.6& \multicolumn{2}{c}{-} \\
TATS~\cite{ge2022long} 3D-VQ & 
162 & 80.6 & \multicolumn{2}{c}{-}& \multicolumn{2}{c}{-} \\ \midrule
& & & \multicolumn{2}{c}{Average} & \multicolumn{2}{c}{Central} \\
\emph{\modelname{}} 3D-VQ-B (ours) & 
127 & 82.1 & 103 & 84.8 & 58 & 87.0 \\ 
\emph{\modelname{}} 3D-VQ-L (ours)
& 45 & 87.1 & 35 & 88.3 & \textbf{25} & \textbf{88.9} \\
\bottomrule
\end{tabular}
\vspace{-1mm}
\caption{\textbf{Comparison of tokenizer architectures and initialization methods} on UCF-101 training set reconstruction results.
The 2D-VQ compresses by 8$\times$8 spatially and the 3D-VQ compresses by 4$\times$8$\times$8 spatial-temporally.
}
\label{tab:token}
\vspace{-4mm}
\end{table}

We evaluate the design options of our 3D-VQ model in \modelname{}.
\cref{tab:token} lists the reconstruction FVD and IS metrics on the UCF-101 training set, which are different from the generation metrics as they measure the intermediate quantization. Nevertheless, reconstruction quality bounds the generation quality.

\cref{tab:token} compares the proposed 3D architecture with existing 2D~\cite{chang2022maskgit} and 3D~\cite{ge2022long} VQ architectures. 
We train the MaskGIT~\cite{chang2022maskgit} 2D-VQ and our 3D-VQ with the same protocol and evaluate the official TATS~\cite{ge2022long} 3D-VQ model.
We compare two inflation methods for our 3D-VQ model, \ie, average~\cite{carreira2017quo} and central inflation.

The results show the following. First, 3D-VQ models, despite producing a higher compression rate, show better video reconstruction quality than 2D-VQ, even with fewer parameters. 
Second, the proposed VQ performs favorably against baseline architectures with a similar size and gets much better with a larger model.
Third, ImageNet~\cite{deng2009imagenet} initialization boosts the performance for 2D and 3D models, where the central inflation outperforms the average inflation.
The results demonstrate the excellent reconstruction fidelity of our tokenizer design.

\section{Related Work}
\label{sec:related}
\vspace{-1mm}
\paragraph{GAN-based approaches.}

Early success in video synthesis has been made by GAN models~\cite{vondrick2016generating, saito2017temporal, tulyakov2018mocogan, clark2019adversarial, brock2018large, yu2022generating,brooks2022generating,skorokhodov2022stylegan,gupta2022rv,acharya2018towards,kahembwe2020lower,tian2021good}. 
Training instability and lack of generation diversity~\cite{chang2022maskgit} are known issues of GAN models.

\vspace{-4mm}
\paragraph{Autoregressive transformers.}
Inspired by the success of GPT~\cite{brown2020language}, autoregressive transformers have been adapted for image~\cite{esser2021taming,ramesh2021zero,yu2022scaling,ding2021cogview,chen2020generative} and video generation~\cite{weissenborn2019scaling,babaeizadeh2021fitvid,wu2021n,hong2022cogvideo}.
A focus for video is autoregressive modeling of visual dynamics. Studies have switched from modeling the raw pixels~\cite{oord2016pixelrnn,chen2020generative} to the discrete codes in a latent space~\cite{rakhimov2021latent,yan2021videogpt}.
The state-of-the-art model TATS~\cite{ge2022long} uses two hierarchical transformers to reduce the computation for long video generation, with tokens learned by a 3D-VQGAN~\cite{esser2021taming}. 
Unlike prior works, we introduce a non-autoregressive transformer with higher efficiency and flexibility.

\vspace{-4mm}
\paragraph{Non-autoregressive transformers.}
Concurrently, a few methods use non-autoregressive transformers for image synthesis~\cite{chang2022maskgit,zhang2021m6,lezama2022improved,sohn2022visual}. \cref{sec:background} reviews a state-of-the-art model called MaskGIT~\cite{chang2022maskgit}. Compared with these approaches~\cite{gupta2022maskvit, han2022show}, we present an embedding mask to model multi-task video conditions with better quality.


\vspace{-4mm}
\paragraph{Diffusion models.}
Diffusion models have recently received much attention for image synthesis.
For example, the state-of-the-art video diffusion model~\cite{ho2022video} extends the image denoising diffusion model~\cite{sohl2015deep,austin2021structured,song2019generative,tzen2019neural,ho2020denoising} by incorporating 3D U-Net~\cite{cciccek20163d} architectures and joint training on both images and videos.
Despite its high-quality, sampling speed is a bottleneck hindering the application of diffusion models in video synthesis. We show a different solution to train a highly-efficient model that offers compelling quality.  

\vspace{-4mm}
\paragraph{Multi-task video synthesis.} 
Multi-task video synthesis~\cite{nash2022transframer, han2022show, wu2021n} is yet to be well-studied.  
Transframer~\cite{nash2022transframer} is the closest to our work, which adopts an image-level representation for autoregressive modeling of tasks based on frame prediction. 
We present an efficient non-autoregressive multi-task transformer, and verify the quality and efficiency on ten video generation tasks.

\vspace{-4mm}
\paragraph{Text-to-video.} All of our models are trained only on public benchmarks, except the Web video model. We leave the text-to-video task as future work. As shown in recent works
~\cite{ho2022imagen,villegas2022phenaki,singer2022make}, training such models requires large, and sometimes non-public, datasets of paired texts and images. 


\section{Conclusion}
\label{sec:conclusion}

In this paper, we propose \emph{\modelname{}}, a generic and efficient mask-based video generation model.
We introduce a high-quality 3D-VQ tokenizer to quantize a video and
design \emph{\methodname{}} for multi-task conditional masked token modeling.
We conduct extensive experiments to demonstrate the video generation quality, efficiency, and flexibility for multi-task generation.
Notably, \modelname{} establishes a new state-of-the-art quality for class conditional generation on UCF-101 and frame prediction on BAIR Robot Pushing and Kinetics-600 datasets.

\clearpage

\iftoggle{cvprpagenumbers}{%
\section*{Acknowledgements}
The authors would like to thank Tom Duerig, Victor Gomes, Paul Natsev along with the Multipod committee for sponsoring the computing resources.
We appreciate valuable feedback and leadership support from David Salesin, Jay Yagnik, Tomas Izo, and Rahul Sukthankar thoughout the project.
Special thanks to Wolfgang Macherey for supporting the project.
We thank David Alexander Ross and Yu-Chuan Su for many helpful comments for improving the paper.
We also give thanks to Sarah Laszlo and Hugh Williams for creating the \modelname{} model card, Bryan Seybold and Albert Shaw for extending the features, Jonathan Ho and Tim Salimans for providing the JAX code pointer for FVD computation, and the Scenic team for the infrastructure support. We are thankful to Wenhe Liu, Xinyu Yao, Mingzhi Cai, Yizhi Zhang, and Zhao Jin for proof reading the paper.}{}

{\small
\bibliographystyle{ieee_fullname}
\bibliography{reference}

\begin{thebibliography}{10}\itemsep=-1pt

\bibitem{acharya2018towards}
Dinesh Acharya, Zhiwu Huang, Danda~Pani Paudel, and Luc Van~Gool.
\newblock Towards high resolution video generation with progressive growing of
  sliced wasserstein gans.
\newblock {\em arXiv:1810.02419}, 2018.

\bibitem{ahmadyan2021objectron}
Adel Ahmadyan, Liangkai Zhang, Artsiom Ablavatski, Jianing Wei, and Matthias
  Grundmann.
\newblock Objectron: A large scale dataset of object-centric videos in the wild
  with pose annotations.
\newblock In {\em CVPR}, 2021.

\bibitem{austin2021structured}
Jacob Austin, Daniel Johnson, Jonathan Ho, Daniel Tarlow, and Rianne van~den
  Berg.
\newblock Structured denoising diffusion models in discrete state-spaces.
\newblock In {\em NeurIPS}, 2021.

\bibitem{babaeizadeh2021fitvid}
Mohammad Babaeizadeh, Mohammad~Taghi Saffar, Suraj Nair, Sergey Levine, Chelsea
  Finn, and Dumitru Erhan.
\newblock Fitvid: Overfitting in pixel-level video prediction.
\newblock {\em arXiv:2106.13195}, 2021.

\bibitem{jax2018github}
James Bradbury, Roy Frostig, Peter Hawkins, Matthew~James Johnson, Chris Leary,
  Dougal Maclaurin, George Necula, Adam Paszke, Jake Vander{P}las, Skye
  Wanderman-{M}ilne, and Qiao Zhang.
\newblock {JAX}: composable transformations of {P}ython+{N}um{P}y programs,
  2018.

\bibitem{brock2018large}
Andrew Brock, Jeff Donahue, and Karen Simonyan.
\newblock Large scale gan training for high fidelity natural image synthesis.
\newblock In {\em ICLR}, 2018.

\bibitem{brooks2022generating}
Tim Brooks, Janne Hellsten, Miika Aittala, Ting-Chun Wang, Timo Aila, Jaakko
  Lehtinen, Ming-Yu Liu, Alexei~A Efros, and Tero Karras.
\newblock Generating long videos of dynamic scenes.
\newblock {\em arXiv:2206.03429}, 2022.

\bibitem{brown2020language}
Tom Brown, Benjamin Mann, Nick Ryder, Melanie Subbiah, Jared~D Kaplan, Prafulla
  Dhariwal, Arvind Neelakantan, Pranav Shyam, Girish Sastry, Amanda Askell,
  et~al.
\newblock Language models are few-shot learners.
\newblock In {\em NeurIPS}, 2020.

\bibitem{caesar2020nuscenes}
Holger Caesar, Varun Bankiti, Alex~H Lang, Sourabh Vora, Venice~Erin Liong,
  Qiang Xu, Anush Krishnan, Yu Pan, Giancarlo Baldan, and Oscar Beijbom.
\newblock nu{S}cenes: A multimodal dataset for autonomous driving.
\newblock In {\em CVPR}, 2020.

\bibitem{carreira2018short}
Joao Carreira, Eric Noland, Andras Banki-Horvath, Chloe Hillier, and Andrew
  Zisserman.
\newblock A short note about {K}inetics-600.
\newblock {\em arXiv:1808.01340}, 2018.

\bibitem{carreira2017quo}
Joao Carreira and Andrew Zisserman.
\newblock Quo vadis, action recognition? a new model and the {K}inetics
  dataset.
\newblock In {\em CVPR}, 2017.

\bibitem{chang2022maskgit}
Huiwen Chang, Han Zhang, Lu Jiang, Ce Liu, and William~T Freeman.
\newblock Mask{GIT}: Masked generative image transformer.
\newblock In {\em CVPR}, 2022.

\bibitem{chen2020generative}
Mark Chen, Alec Radford, Rewon Child, Jeffrey Wu, Heewoo Jun, David Luan, and
  Ilya Sutskever.
\newblock Generative pretraining from pixels.
\newblock In {\em ICML}, 2020.

\bibitem{cciccek20163d}
{\"O}zg{\"u}n {\c{C}}i{\c{c}}ek, Ahmed Abdulkadir, Soeren~S Lienkamp, Thomas
  Brox, and Olaf Ronneberger.
\newblock 3{D} {U}-{N}et: learning dense volumetric segmentation from sparse
  annotation.
\newblock In {\em MICCAI}, 2016.

\bibitem{clark2019adversarial}
Aidan Clark, Jeff Donahue, and Karen Simonyan.
\newblock Adversarial video generation on complex datasets.
\newblock {\em arXiv:1907.06571}, 2019.

\bibitem{deng2009imagenet}
Jia Deng, Wei Dong, Richard Socher, Li-Jia Li, Kai Li, and Li Fei-Fei.
\newblock Image{N}et: A large-scale hierarchical image database.
\newblock In {\em CVPR}, 2009.

\bibitem{devlin2019bert}
Jacob Devlin, Ming-Wei Chang, Kenton Lee, and Kristina Toutanova.
\newblock {BERT}: Pre-training of deep bidirectional transformers for language
  understanding.
\newblock In {\em NAACL}, 2019.

\bibitem{ding2021cogview}
Ming Ding, Zhuoyi Yang, Wenyi Hong, Wendi Zheng, Chang Zhou, Da Yin, Junyang
  Lin, Xu Zou, Zhou Shao, Hongxia Yang, et~al.
\newblock Cog{V}iew: Mastering text-to-image generation via transformers.
\newblock In {\em NeurIPS}, 2021.

\bibitem{ebert2017self}
Frederik Ebert, Chelsea Finn, Alex~X Lee, and Sergey Levine.
\newblock Self-supervised visual planning with temporal skip connections.
\newblock In {\em CoRL}, 2017.

\bibitem{esser2021taming}
Patrick Esser, Robin Rombach, and Bjorn Ommer.
\newblock Taming transformers for high-resolution image synthesis.
\newblock In {\em CVPR}, 2021.

\bibitem{ge2022long}
Songwei Ge, Thomas Hayes, Harry Yang, Xi Yin, Guan Pang, David Jacobs, Jia-Bin
  Huang, and Devi Parikh.
\newblock Long video generation with time-agnostic {VQGAN} and time-sensitive
  transformer.
\newblock In {\em ECCV}, 2022.

\bibitem{ghazvininejad2019mask}
Marjan Ghazvininejad, Omer Levy, Yinhan Liu, and Luke Zettlemoyer.
\newblock Mask-{P}redict: Parallel decoding of conditional masked language
  models.
\newblock In {\em EMNLP-IJCNLP}, 2019.

\bibitem{goyal2017something}
Raghav Goyal, Samira Ebrahimi~Kahou, Vincent Michalski, Joanna Materzynska,
  Susanne Westphal, Heuna Kim, Valentin Haenel, Ingo Fruend, Peter Yianilos,
  Moritz Mueller-Freitag, et~al.
\newblock The ``something something" video database for learning and evaluating
  visual common sense.
\newblock In {\em ICCV}, 2017.

\bibitem{gu2021fully}
Jiatao Gu and Xiang Kong.
\newblock Fully non-autoregressive neural machine translation: Tricks of the
  trade.
\newblock In {\em ACL-IJCNLP Findings}, 2021.

\bibitem{gu2022vector}
Shuyang Gu, Dong Chen, Jianmin Bao, Fang Wen, Bo Zhang, Dongdong Chen, Lu Yuan,
  and Baining Guo.
\newblock Vector quantized diffusion model for text-to-image synthesis.
\newblock In {\em CVPR}, 2022.

\bibitem{gupta2022maskvit}
Agrim Gupta, Stephen Tian, Yunzhi Zhang, Jiajun Wu, Roberto
  Mart{\'\i}n-Mart{\'\i}n, and Li Fei-Fei.
\newblock Mask{V}i{T}: Masked visual pre-training for video prediction.
\newblock {\em arXiv:2206.11894}, 2022.

\bibitem{gupta2022rv}
Sonam Gupta, Arti Keshari, and Sukhendu Das.
\newblock {RV-GAN}: Recurrent {GAN} for unconditional video generation.
\newblock In {\em CVPRW}, 2022.

\bibitem{han2022show}
Ligong Han, Jian Ren, Hsin-Ying Lee, Francesco Barbieri, Kyle Olszewski,
  Shervin Minaee, Dimitris Metaxas, and Sergey Tulyakov.
\newblock Show me what and tell me how: Video synthesis via multimodal
  conditioning.
\newblock In {\em CVPR}, 2022.

\bibitem{he2016deep}
Kaiming He, Xiangyu Zhang, Shaoqing Ren, and Jian Sun.
\newblock Deep residual learning for image recognition.
\newblock In {\em CVPR}, 2016.

\bibitem{flax2020github}
Jonathan Heek, Anselm Levskaya, Avital Oliver, Marvin Ritter, Bertrand
  Rondepierre, Andreas Steiner, and Marc van {Z}ee.
\newblock {F}lax: A neural network library and ecosystem for {JAX}, 2020.

\bibitem{ho2022imagen}
Jonathan Ho, William Chan, Chitwan Saharia, Jay Whang, Ruiqi Gao, Alexey
  Gritsenko, Diederik~P Kingma, Ben Poole, Mohammad Norouzi, David~J Fleet,
  et~al.
\newblock Imagen video: High definition video generation with diffusion models.
\newblock {\em arXiv:2210.02303}, 2022.

\bibitem{ho2020denoising}
Jonathan Ho, Ajay Jain, and Pieter Abbeel.
\newblock Denoising diffusion probabilistic models.
\newblock In {\em NeurIPS}, 2020.

\bibitem{ho2022video}
Jonathan Ho, Tim Salimans, Alexey Gritsenko, William Chan, Mohammad Norouzi,
  and David~J Fleet.
\newblock Video diffusion models.
\newblock In {\em ICLR Workshops}, 2022.

\bibitem{hong2022cogvideo}
Wenyi Hong, Ming Ding, Wendi Zheng, Xinghan Liu, and Jie Tang.
\newblock Cog{V}ideo: Large-scale pretraining for text-to-video generation via
  transformers.
\newblock {\em arXiv:2205.15868}, 2022.

\bibitem{hoppe2022diffusion}
Tobias H{\"o}ppe, Arash Mehrjou, Stefan Bauer, Didrik Nielsen, and Andrea
  Dittadi.
\newblock Diffusion models for video prediction and infilling.
\newblock {\em arXiv:2206.07696}, 2022.

\bibitem{jouppi2021ten}
Norman~P Jouppi, Doe~Hyun Yoon, Matthew Ashcraft, Mark Gottscho, Thomas~B
  Jablin, George Kurian, James Laudon, Sheng Li, Peter Ma, Xiaoyu Ma, et~al.
\newblock Ten lessons from three generations shaped google’s {TPUv4i}.
\newblock In {\em ISCA}, 2021.

\bibitem{kahembwe2020lower}
Emmanuel Kahembwe and Subramanian Ramamoorthy.
\newblock Lower dimensional kernels for video discriminators.
\newblock {\em Neural Networks}, 132:506--520, 2020.

\bibitem{karras2019style}
Tero Karras, Samuli Laine, and Timo Aila.
\newblock A style-based generator architecture for generative adversarial
  networks.
\newblock In {\em CVPR}, 2019.

\bibitem{kingma2014adam}
Diederik~P Kingma and Jimmy Ba.
\newblock Adam: A method for stochastic optimization.
\newblock {\em arXiv:1412.6980}, 2014.

\bibitem{kong2021blt}
Xiang Kong, Lu Jiang, Huiwen Chang, Han Zhang, Yuan Hao, Haifeng Gong, and
  Irfan Essa.
\newblock {BLT}: Bidirectional layout transformer for controllable layout
  generation.
\newblock In {\em ECCV}, 2022.

\bibitem{le2021ccvs}
Guillaume Le~Moing, Jean Ponce, and Cordelia Schmid.
\newblock {CCVS}: Context-aware controllable video synthesis.
\newblock In {\em NeurIPS}, 2021.

\bibitem{lezama2022improved}
Jos{\'e} Lezama, Huiwen Chang, Lu Jiang, and Irfan Essa.
\newblock Improved masked image generation with {T}oken-{C}ritic.
\newblock In {\em ECCV}, 2022.

\bibitem{luc2020transformation}
Pauline Luc, Aidan Clark, Sander Dieleman, Diego de~Las Casas, Yotam Doron,
  Albin Cassirer, and Karen Simonyan.
\newblock Transformation-based adversarial video prediction on large-scale
  data.
\newblock {\em arXiv:2003.04035}, 2020.

\bibitem{nash2022transframer}
Charlie Nash, Jo{\~a}o Carreira, Jacob Walker, Iain Barr, Andrew Jaegle,
  Mateusz Malinowski, and Peter Battaglia.
\newblock Transframer: Arbitrary frame prediction with generative models.
\newblock {\em arXiv:2203.09494}, 2022.

\bibitem{rakhimov2021latent}
Ruslan Rakhimov, Denis Volkhonskiy, Alexey Artemov, Denis Zorin, and Evgeny
  Burnaev.
\newblock Latent video transformer.
\newblock In {\em VISIGRAPP (5: VISAPP)}, 2021.

\bibitem{ramesh2021zero}
Aditya Ramesh, Mikhail Pavlov, Gabriel Goh, Scott Gray, Chelsea Voss, Alec
  Radford, Mark Chen, and Ilya Sutskever.
\newblock Zero-shot text-to-image generation.
\newblock In {\em ICML}, 2021.

\bibitem{rombach2022high}
Robin Rombach, Andreas Blattmann, Dominik Lorenz, Patrick Esser, and Bj{\"o}rn
  Ommer.
\newblock High-resolution image synthesis with latent diffusion models.
\newblock In {\em CVPR}, 2022.

\bibitem{saito2017temporal}
Masaki Saito, Eiichi Matsumoto, and Shunta Saito.
\newblock Temporal generative adversarial nets with singular value clipping.
\newblock In {\em ICCV}, 2017.

\bibitem{saito2020train}
Masaki Saito, Shunta Saito, Masanori Koyama, and Sosuke Kobayashi.
\newblock Train sparsely, generate densely: Memory-efficient unsupervised
  training of high-resolution temporal gan.
\newblock {\em IJCV}, 128(10):2586--2606, 2020.

\bibitem{singer2022make}
Uriel Singer, Adam Polyak, Thomas Hayes, Xi Yin, Jie An, Songyang Zhang, Qiyuan
  Hu, Harry Yang, Oron Ashual, Oran Gafni, et~al.
\newblock Make-a-video: Text-to-video generation without text-video data.
\newblock {\em arXiv:2209.14792}, 2022.

\bibitem{skorokhodov2022stylegan}
Ivan Skorokhodov, Sergey Tulyakov, and Mohamed Elhoseiny.
\newblock {StyleGAN-V}: A continuous video generator with the price, image
  quality and perks of {StyleGAN2}.
\newblock In {\em CVPR}, 2022.

\bibitem{sohl2015deep}
Jascha Sohl-Dickstein, Eric Weiss, Niru Maheswaranathan, and Surya Ganguli.
\newblock Deep unsupervised learning using nonequilibrium thermodynamics.
\newblock In {\em ICML}, 2015.

\bibitem{song2019generative}
Yang Song and Stefano Ermon.
\newblock Generative modeling by estimating gradients of the data distribution.
\newblock In {\em NeurIPS}, 2019.

\bibitem{soomro2012ucf101}
Khurram Soomro, Amir~Roshan Zamir, and Mubarak Shah.
\newblock {UCF101}: A dataset of 101 human actions classes from videos in the
  wild.
\newblock {\em arXiv:1212.0402}, 2012.

\bibitem{tian2021good}
Yu Tian, Jian Ren, Menglei Chai, Kyle Olszewski, Xi Peng, Dimitris~N Metaxas,
  and Sergey Tulyakov.
\newblock A good image generator is what you need for high-resolution video
  synthesis.
\newblock In {\em ICLR}, 2021.

\bibitem{tran2015learning}
Du Tran, Lubomir Bourdev, Rob Fergus, Lorenzo Torresani, and Manohar Paluri.
\newblock Learning spatiotemporal features with 3{D} convolutional networks.
\newblock In {\em ICCV}, 2015.

\bibitem{tseng2021regularizing}
Hung-Yu Tseng, Lu Jiang, Ce Liu, Ming-Hsuan Yang, and Weilong Yang.
\newblock Regularizing generative adversarial networks under limited data.
\newblock In {\em CVPR}, 2021.

\bibitem{tulyakov2018mocogan}
Sergey Tulyakov, Ming-Yu Liu, Xiaodong Yang, and Jan Kautz.
\newblock Mo{C}o{GAN}: Decomposing motion and content for video generation.
\newblock In {\em CVPR}, 2018.

\bibitem{tzen2019neural}
Belinda Tzen and Maxim Raginsky.
\newblock Neural stochastic differential equations: Deep latent gaussian models
  in the diffusion limit.
\newblock {\em arXiv:1905.09883}, 2019.

\bibitem{unterthiner2018towards}
Thomas Unterthiner, Sjoerd van Steenkiste, Karol Kurach, Raphael Marinier,
  Marcin Michalski, and Sylvain Gelly.
\newblock Towards accurate generative models of video: A new metric \&
  challenges.
\newblock {\em arXiv:1812.01717}, 2018.

\bibitem{oord2016pixelrnn}
A{\"{a}}ron van~den Oord, Nal Kalchbrenner, and Koray Kavukcuoglu.
\newblock Pixel recurrent neural networks.
\newblock In {\em ICML}, 2016.

\bibitem{van2017neural}
Aaron Van Den~Oord, Oriol Vinyals, et~al.
\newblock Neural discrete representation learning.
\newblock In {\em NeurIPS}, 2017.

\bibitem{villegas2022phenaki}
Ruben Villegas, Mohammad Babaeizadeh, Pieter-Jan Kindermans, Hernan Moraldo,
  Han Zhang, Mohammad~Taghi Saffar, Santiago Castro, Julius Kunze, and Dumitru
  Erhan.
\newblock Phenaki: Variable length video generation from open domain textual
  description.
\newblock {\em arXiv:2210.02399}, 2022.

\bibitem{voleti2022masked}
Vikram Voleti, Alexia Jolicoeur-Martineau, and Christopher Pal.
\newblock Masked conditional video diffusion for prediction, generation, and
  interpolation.
\newblock In {\em NeurIPS}, 2022.

\bibitem{vondrick2016generating}
Carl Vondrick, Hamed Pirsiavash, and Antonio Torralba.
\newblock Generating videos with scene dynamics.
\newblock In {\em NeurIPS}, 2016.

\bibitem{wang2022image}
Wenhui Wang, Hangbo Bao, Li Dong, Johan Bjorck, Zhiliang Peng, Qiang Liu, Kriti
  Aggarwal, Owais~Khan Mohammed, Saksham Singhal, Subhojit Som, et~al.
\newblock Image as a foreign language: {BEiT} pretraining for all vision and
  vision-language tasks.
\newblock {\em arXiv:2208.10442}, 2022.

\bibitem{wang2004image}
Zhou Wang, Alan~C Bovik, Hamid~R Sheikh, and Eero~P Simoncelli.
\newblock Image quality assessment: from error visibility to structural
  similarity.
\newblock {\em IEEE TIP}, 13(4):600--612, 2004.

\bibitem{weissenborn2019scaling}
Dirk Weissenborn, Oscar T{\"a}ckstr{\"o}m, and Jakob Uszkoreit.
\newblock Scaling autoregressive video models.
\newblock In {\em ICLR}, 2019.

\bibitem{wu2021n}
Chenfei Wu, Jian Liang, Lei Ji, Fan Yang, Yuejian Fang, Daxin Jiang, and Nan
  Duan.
\newblock {N\"UWA}: Visual synthesis pre-training for neural visual world
  creation.
\newblock In {\em ECCV}, 2022.

\bibitem{yan2021videogpt}
Wilson Yan, Yunzhi Zhang, Pieter Abbeel, and Aravind Srinivas.
\newblock Video{GPT}: Video generation using vq-vae and transformers.
\newblock {\em arXiv:2104.10157}, 2021.

\bibitem{yu2021vector}
Jiahui Yu, Xin Li, Jing~Yu Koh, Han Zhang, Ruoming Pang, James Qin, Alexander
  Ku, Yuanzhong Xu, Jason Baldridge, and Yonghui Wu.
\newblock Vector-quantized image modeling with improved {VQGAN}.
\newblock In {\em ICLR}, 2022.

\bibitem{yu2022scaling}
Jiahui Yu, Yuanzhong Xu, Jing~Yu Koh, Thang Luong, Gunjan Baid, Zirui Wang,
  Vijay Vasudevan, Alexander Ku, Yinfei Yang, Burcu~Karagol Ayan, et~al.
\newblock Scaling autoregressive models for content-rich text-to-image
  generation.
\newblock {\em arXiv:2206.10789}, 2022.

\bibitem{yu2022generating}
Sihyun Yu, Jihoon Tack, Sangwoo Mo, Hyunsu Kim, Junho Kim, Jung-Woo Ha, and
  Jinwoo Shin.
\newblock Generating videos with dynamics-aware implicit generative adversarial
  networks.
\newblock In {\em ICLR}, 2022.

\bibitem{zhang2018unreasonable}
Richard Zhang, Phillip Isola, Alexei~A Efros, Eli Shechtman, and Oliver Wang.
\newblock The unreasonable effectiveness of deep features as a perceptual
  metric.
\newblock In {\em CVPR}, 2018.

\bibitem{zhang2021m6}
Zhu Zhang, Jianxin Ma, Chang Zhou, Rui Men, Zhikang Li, Ming Ding, Jie Tang,
  Jingren Zhou, and Hongxia Yang.
\newblock M6-{UFC}: Unifying multi-modal controls for conditional image
  synthesis.
\newblock {\em arXiv:2105.14211}, 2021.

\end{thebibliography}


\begin{thebibliography}{10}\itemsep=-1pt

\bibitem{abu2016youtube}
Sami Abu-El-Haija, Nisarg Kothari, Joonseok Lee, Paul Natsev, George Toderici,
  Balakrishnan Varadarajan, and Sudheendra Vijayanarasimhan.
\newblock Youtube-8m: A large-scale video classification benchmark.
\newblock {\em arXiv preprint arXiv:1609.08675}, 2016.

\bibitem{acharya2018towards}
Dinesh Acharya, Zhiwu Huang, Danda~Pani Paudel, and Luc Van~Gool.
\newblock Towards high resolution video generation with progressive growing of
  sliced wasserstein gans.
\newblock {\em arXiv:1810.02419}, 2018.

\bibitem{ahmadyan2021objectron}
Adel Ahmadyan, Liangkai Zhang, Artsiom Ablavatski, Jianing Wei, and Matthias
  Grundmann.
\newblock Objectron: A large scale dataset of object-centric videos in the wild
  with pose annotations.
\newblock In {\em CVPR}, 2021.

\bibitem{babaeizadeh2021fitvid}
Mohammad Babaeizadeh, Mohammad~Taghi Saffar, Suraj Nair, Sergey Levine, Chelsea
  Finn, and Dumitru Erhan.
\newblock Fitvid: Overfitting in pixel-level video prediction.
\newblock {\em arXiv:2106.13195}, 2021.

\bibitem{caesar2020nuscenes}
Holger Caesar, Varun Bankiti, Alex~H Lang, Sourabh Vora, Venice~Erin Liong,
  Qiang Xu, Anush Krishnan, Yu Pan, Giancarlo Baldan, and Oscar Beijbom.
\newblock nu{S}cenes: A multimodal dataset for autonomous driving.
\newblock In {\em CVPR}, 2020.

\bibitem{carreira2018short}
Joao Carreira, Eric Noland, Andras Banki-Horvath, Chloe Hillier, and Andrew
  Zisserman.
\newblock A short note about {K}inetics-600.
\newblock {\em arXiv:1808.01340}, 2018.

\bibitem{carreira2017quo}
Joao Carreira and Andrew Zisserman.
\newblock Quo vadis, action recognition? a new model and the {K}inetics
  dataset.
\newblock In {\em CVPR}, 2017.

\bibitem{chang2022maskgit}
Huiwen Chang, Han Zhang, Lu Jiang, Ce Liu, and William~T Freeman.
\newblock Mask{GIT}: Masked generative image transformer.
\newblock In {\em CVPR}, 2022.

\bibitem{clark2019adversarial}
Aidan Clark, Jeff Donahue, and Karen Simonyan.
\newblock Adversarial video generation on complex datasets.
\newblock {\em arXiv:1907.06571}, 2019.

\bibitem{devlin2019bert}
Jacob Devlin, Ming-Wei Chang, Kenton Lee, and Kristina Toutanova.
\newblock {BERT}: Pre-training of deep bidirectional transformers for language
  understanding.
\newblock In {\em NAACL}, 2019.

\bibitem{dosovitskiy2020image}
Alexey Dosovitskiy, Lucas Beyer, Alexander Kolesnikov, Dirk Weissenborn,
  Xiaohua Zhai, Thomas Unterthiner, Mostafa Dehghani, Matthias Minderer, Georg
  Heigold, Sylvain Gelly, et~al.
\newblock An image is worth 16x16 words: Transformers for image recognition at
  scale.
\newblock In {\em ICLR}, 2020.

\bibitem{ebert2017self}
Frederik Ebert, Chelsea Finn, Alex~X Lee, and Sergey Levine.
\newblock Self-supervised visual planning with temporal skip connections.
\newblock In {\em CoRL}, 2017.

\bibitem{ge2022long}
Songwei Ge, Thomas Hayes, Harry Yang, Xi Yin, Guan Pang, David Jacobs, Jia-Bin
  Huang, and Devi Parikh.
\newblock Long video generation with time-agnostic {VQGAN} and time-sensitive
  transformer.
\newblock In {\em ECCV}, 2022.

\bibitem{goyal2017something}
Raghav Goyal, Samira Ebrahimi~Kahou, Vincent Michalski, Joanna Materzynska,
  Susanne Westphal, Heuna Kim, Valentin Haenel, Ingo Fruend, Peter Yianilos,
  Moritz Mueller-Freitag, et~al.
\newblock The ``something something" video database for learning and evaluating
  visual common sense.
\newblock In {\em ICCV}, 2017.

\bibitem{gupta2022maskvit}
Agrim Gupta, Stephen Tian, Yunzhi Zhang, Jiajun Wu, Roberto
  Mart{\'\i}n-Mart{\'\i}n, and Li Fei-Fei.
\newblock Mask{V}i{T}: Masked visual pre-training for video prediction.
\newblock {\em arXiv:2206.11894}, 2022.

\bibitem{hendrycks2016gaussian}
Dan Hendrycks and Kevin Gimpel.
\newblock Gaussian error linear units (gelus).
\newblock {\em arXiv:1606.08415}, 2016.

\bibitem{ho2022video}
Jonathan Ho, Tim Salimans, Alexey Gritsenko, William Chan, Mohammad Norouzi,
  and David~J Fleet.
\newblock Video diffusion models.
\newblock In {\em ICLR Workshops}, 2022.

\bibitem{hong2022cogvideo}
Wenyi Hong, Ming Ding, Wendi Zheng, Xinghan Liu, and Jie Tang.
\newblock Cog{V}ideo: Large-scale pretraining for text-to-video generation via
  transformers.
\newblock {\em arXiv:2205.15868}, 2022.

\bibitem{hoppe2022diffusion}
Tobias H{\"o}ppe, Arash Mehrjou, Stefan Bauer, Didrik Nielsen, and Andrea
  Dittadi.
\newblock Diffusion models for video prediction and infilling.
\newblock {\em arXiv:2206.07696}, 2022.

\bibitem{ioffe2015batch}
Sergey Ioffe and Christian Szegedy.
\newblock Batch normalization: Accelerating deep network training by reducing
  internal covariate shift.
\newblock In {\em ICML}, 2015.

\bibitem{kahembwe2020lower}
Emmanuel Kahembwe and Subramanian Ramamoorthy.
\newblock Lower dimensional kernels for video discriminators.
\newblock {\em Neural Networks}, 132:506--520, 2020.

\bibitem{le2021ccvs}
Guillaume Le~Moing, Jean Ponce, and Cordelia Schmid.
\newblock {CCVS}: Context-aware controllable video synthesis.
\newblock In {\em NeurIPS}, 2021.

\bibitem{luc2020transformation}
Pauline Luc, Aidan Clark, Sander Dieleman, Diego de~Las Casas, Yotam Doron,
  Albin Cassirer, and Karen Simonyan.
\newblock Transformation-based adversarial video prediction on large-scale
  data.
\newblock {\em arXiv:2003.04035}, 2020.

\bibitem{nash2022transframer}
Charlie Nash, Jo{\~a}o Carreira, Jacob Walker, Iain Barr, Andrew Jaegle,
  Mateusz Malinowski, and Peter Battaglia.
\newblock Transframer: Arbitrary frame prediction with generative models.
\newblock {\em arXiv:2203.09494}, 2022.

\bibitem{rakhimov2021latent}
Ruslan Rakhimov, Denis Volkhonskiy, Alexey Artemov, Denis Zorin, and Evgeny
  Burnaev.
\newblock Latent video transformer.
\newblock In {\em VISIGRAPP (5: VISAPP)}, 2021.

\bibitem{ramachandran2018searching}
Prajit Ramachandran, Barret Zoph, and Quoc~V Le.
\newblock Searching for activation functions.
\newblock In {\em ICLR Workshops}, 2018.

\bibitem{saito2017temporal}
Masaki Saito, Eiichi Matsumoto, and Shunta Saito.
\newblock Temporal generative adversarial nets with singular value clipping.
\newblock In {\em ICCV}, 2017.

\bibitem{saito2020train}
Masaki Saito, Shunta Saito, Masanori Koyama, and Sosuke Kobayashi.
\newblock Train sparsely, generate densely: Memory-efficient unsupervised
  training of high-resolution temporal gan.
\newblock {\em IJCV}, 128(10):2586--2606, 2020.

\bibitem{singer2022make}
Uriel Singer, Adam Polyak, Thomas Hayes, Xi Yin, Jie An, Songyang Zhang, Qiyuan
  Hu, Harry Yang, Oron Ashual, Oran Gafni, et~al.
\newblock Make-a-video: Text-to-video generation without text-video data.
\newblock {\em arXiv:2209.14792}, 2022.

\bibitem{skorokhodov2022stylegan}
Ivan Skorokhodov, Sergey Tulyakov, and Mohamed Elhoseiny.
\newblock {StyleGAN-V}: A continuous video generator with the price, image
  quality and perks of {StyleGAN2}.
\newblock In {\em CVPR}, 2022.

\bibitem{soomro2012ucf101}
Khurram Soomro, Amir~Roshan Zamir, and Mubarak Shah.
\newblock {UCF101}: A dataset of 101 human actions classes from videos in the
  wild.
\newblock {\em arXiv:1212.0402}, 2012.

\bibitem{tian2021good}
Yu Tian, Jian Ren, Menglei Chai, Kyle Olszewski, Xi Peng, Dimitris~N Metaxas,
  and Sergey Tulyakov.
\newblock A good image generator is what you need for high-resolution video
  synthesis.
\newblock In {\em ICLR}, 2021.

\bibitem{tran2015learning}
Du Tran, Lubomir Bourdev, Rob Fergus, Lorenzo Torresani, and Manohar Paluri.
\newblock Learning spatiotemporal features with 3{D} convolutional networks.
\newblock In {\em ICCV}, 2015.

\bibitem{tulyakov2018mocogan}
Sergey Tulyakov, Ming-Yu Liu, Xiaodong Yang, and Jan Kautz.
\newblock Mo{C}o{GAN}: Decomposing motion and content for video generation.
\newblock In {\em CVPR}, 2018.

\bibitem{unterthiner2018towards}
Thomas Unterthiner, Sjoerd van Steenkiste, Karol Kurach, Raphael Marinier,
  Marcin Michalski, and Sylvain Gelly.
\newblock Towards accurate generative models of video: A new metric \&
  challenges.
\newblock {\em arXiv:1812.01717}, 2018.

\bibitem{villegas2022phenaki}
Ruben Villegas, Mohammad Babaeizadeh, Pieter-Jan Kindermans, Hernan Moraldo,
  Han Zhang, Mohammad~Taghi Saffar, Santiago Castro, Julius Kunze, and Dumitru
  Erhan.
\newblock Phenaki: Variable length video generation from open domain textual
  description.
\newblock {\em arXiv:2210.02399}, 2022.

\bibitem{voleti2022masked}
Vikram Voleti, Alexia Jolicoeur-Martineau, and Christopher Pal.
\newblock Masked conditional video diffusion for prediction, generation, and
  interpolation.
\newblock In {\em NeurIPS}, 2022.

\bibitem{vondrick2016generating}
Carl Vondrick, Hamed Pirsiavash, and Antonio Torralba.
\newblock Generating videos with scene dynamics.
\newblock In {\em NeurIPS}, 2016.

\bibitem{wang2004image}
Zhou Wang, Alan~C Bovik, Hamid~R Sheikh, and Eero~P Simoncelli.
\newblock Image quality assessment: from error visibility to structural
  similarity.
\newblock {\em IEEE TIP}, 13(4):600--612, 2004.

\bibitem{weissenborn2019scaling}
Dirk Weissenborn, Oscar T{\"a}ckstr{\"o}m, and Jakob Uszkoreit.
\newblock Scaling autoregressive video models.
\newblock In {\em ICLR}, 2019.

\bibitem{wu2021n}
Chenfei Wu, Jian Liang, Lei Ji, Fan Yang, Yuejian Fang, Daxin Jiang, and Nan
  Duan.
\newblock {N\"UWA}: Visual synthesis pre-training for neural visual world
  creation.
\newblock In {\em ECCV}, 2022.

\bibitem{wu2018group}
Yuxin Wu and Kaiming He.
\newblock Group normalization.
\newblock In {\em ECCV}, 2018.

\bibitem{yan2021videogpt}
Wilson Yan, Yunzhi Zhang, Pieter Abbeel, and Aravind Srinivas.
\newblock Video{GPT}: Video generation using vq-vae and transformers.
\newblock {\em arXiv:2104.10157}, 2021.

\bibitem{yu2022generating}
Sihyun Yu, Jihoon Tack, Sangwoo Mo, Hyunsu Kim, Junho Kim, Jung-Woo Ha, and
  Jinwoo Shin.
\newblock Generating videos with dynamics-aware implicit generative adversarial
  networks.
\newblock In {\em ICLR}, 2022.

\bibitem{zhang2018unreasonable}
Richard Zhang, Phillip Isola, Alexei~A Efros, Eli Shechtman, and Oliver Wang.
\newblock The unreasonable effectiveness of deep features as a perceptual
  metric.
\newblock In {\em CVPR}, 2018.

\end{thebibliography}
\balance
}

\end{document}


\makeatletter

\newcommand{\lu}[1]{{\textcolor{red}{[lu: #1]}}}
\newcommand{\lijun}[1]{{\textcolor{orange}{[lijun: #1]}}}

\DeclareRobustCommand\onedot{\futurelet\@let@token\@onedot}
\def\@onedot{\ifx\@let@token.\else.\null\fi\xspace}

\def\eg{\emph{e.g}\onedot} \def\Eg{\emph{E.g}\onedot}
\def\ie{\emph{i.e}\onedot} \def\Ie{\emph{I.e}\onedot}
\def\cf{\emph{c.f}\onedot} \def\Cf{\emph{C.f}\onedot}
\def\etc{\emph{etc}\onedot} \def\vs{\emph{vs}\onedot}
\def\wrt{w.r.t\onedot} \def\dof{d.o.f\onedot}
\def\etal{\emph{et al}\onedot}
\makeatother

\newcommand{\modelname}{MAGVIT}
\newcommand{\methodname}{COMMIT}
\newcommand{\mask}{\texttt{[MASK]}\xspace}
\newcommand{\keep}{0\xspace}
\newcommand{\intecond}{\texttt{[INCO]}\xspace}
\newcommand{\pad}{\texttt{[PAD]}\xspace}

\title{\modelname: Masked Generative Video Transformer \\Supplementary Materials}

\author{}
\maketitle

\appendix

\section*{Acknowledgements}
The authors would like to thank Tom Duerig, Victor Gomes, Paul Natsev along with the Multipod committee for sponsoring the computing resources.
We appreciate valuable feedback and leadership support from David Salesin, Jay Yagnik, Tomas Izo, and Rahul Sukthankar thoughout the project.
Special thanks to Wolfgang Macherey for supporting the project.
We thank David Alexander Ross and Yu-Chuan Su for many helpful comments for improving the paper.
We also give thanks to Sarah Laszlo and Hugh Williams for creating the \modelname{} model card, Bryan Seybold and Albert Shaw for extending the features, Jonathan Ho and Tim Salimans for providing the JAX code pointer for FVD computation, and the Scenic team for the infrastructure support. We are thankful to Wenhe Liu, Xinyu Yao, Mingzhi Cai, Yizhi Zhang, and Zhao Jin for proof reading the paper.
This project is funded in part by Carnegie Mellon University’s Mobility21 National University Transportation Center, which is sponsored by the US Department of Transportation.

\section*{Appendix Overview}
This supplementary document provides additional details to support our main manuscript, organized as follows:
\begin{itemize}[nosep, leftmargin=*]
    \item \cref{app:arch} presents the 3D-VQ architectures and the transformer models in \modelname{}.
    \item \cref{app:imp_details} includes additional implementation details in training and evaluation.
    \item \cref{app:quant} provides more quantitative evaluation results, which include:
    \begin{itemize}[nosep,leftmargin=*]
        \item Comparisons to more published results on the three benchmarks in the paper: UCF-101~\cite{soomro2012ucf101}, BAIR~\cite{ebert2017self,unterthiner2018towards}, and Kinetics-600~\cite{carreira2018short}.
        \item Multi-task results on Something-Something-v2 (SSv2)~\cite{goyal2017something}.
        \item Results on three additional datasets: NuScenes~\cite{caesar2020nuscenes}, Objectron~\cite{ahmadyan2021objectron} and Web video datasets.
    \end{itemize}
    \item \cref{app:qualitative} shows more qualitative examples of the generated videos.
\end{itemize}

We present a demo video for \modelname{} and
show more generated examples on this 
\iftoggle{cvprfinal}{%
web page\footnote{\url{https://magvit.cs.cmu.edu}}
}{%
anonymous page ({\small \url{https://magvit-ann.github.io}})
}.
%

\section{\modelname{} Model Architecture}
\label{app:arch}

\begin{figure*}[tp]
\centering
\begin{subfigure}[t]{0.490\linewidth}
\vskip 0pt 
\includegraphics[width=\linewidth]{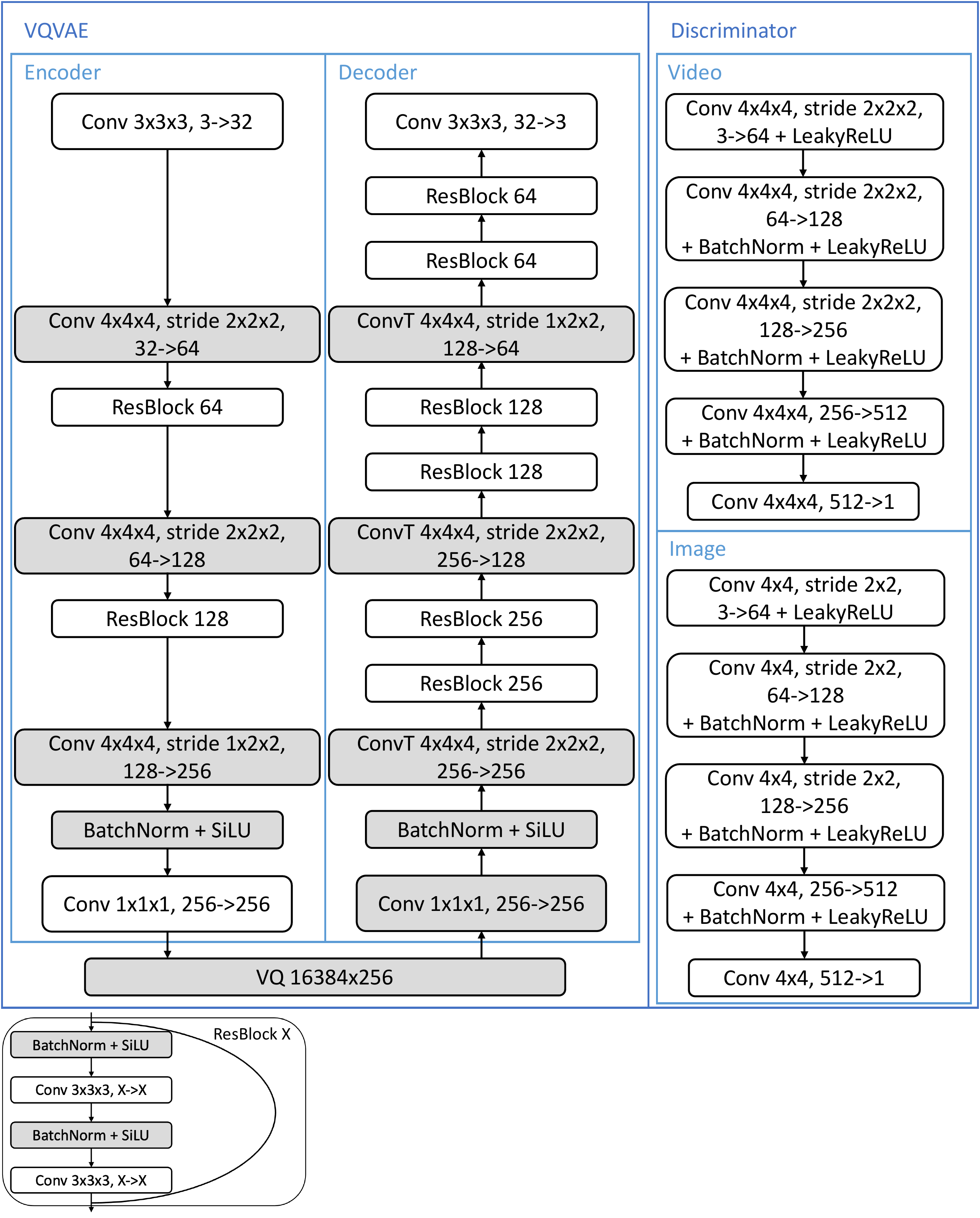}
\caption{TATS~\cite{ge2022long} 3D-VQ. (32M+14M parameters)}
\end{subfigure}
\begin{subfigure}[t]{0.505\linewidth}
\vskip 0pt 
\includegraphics[width=\linewidth]{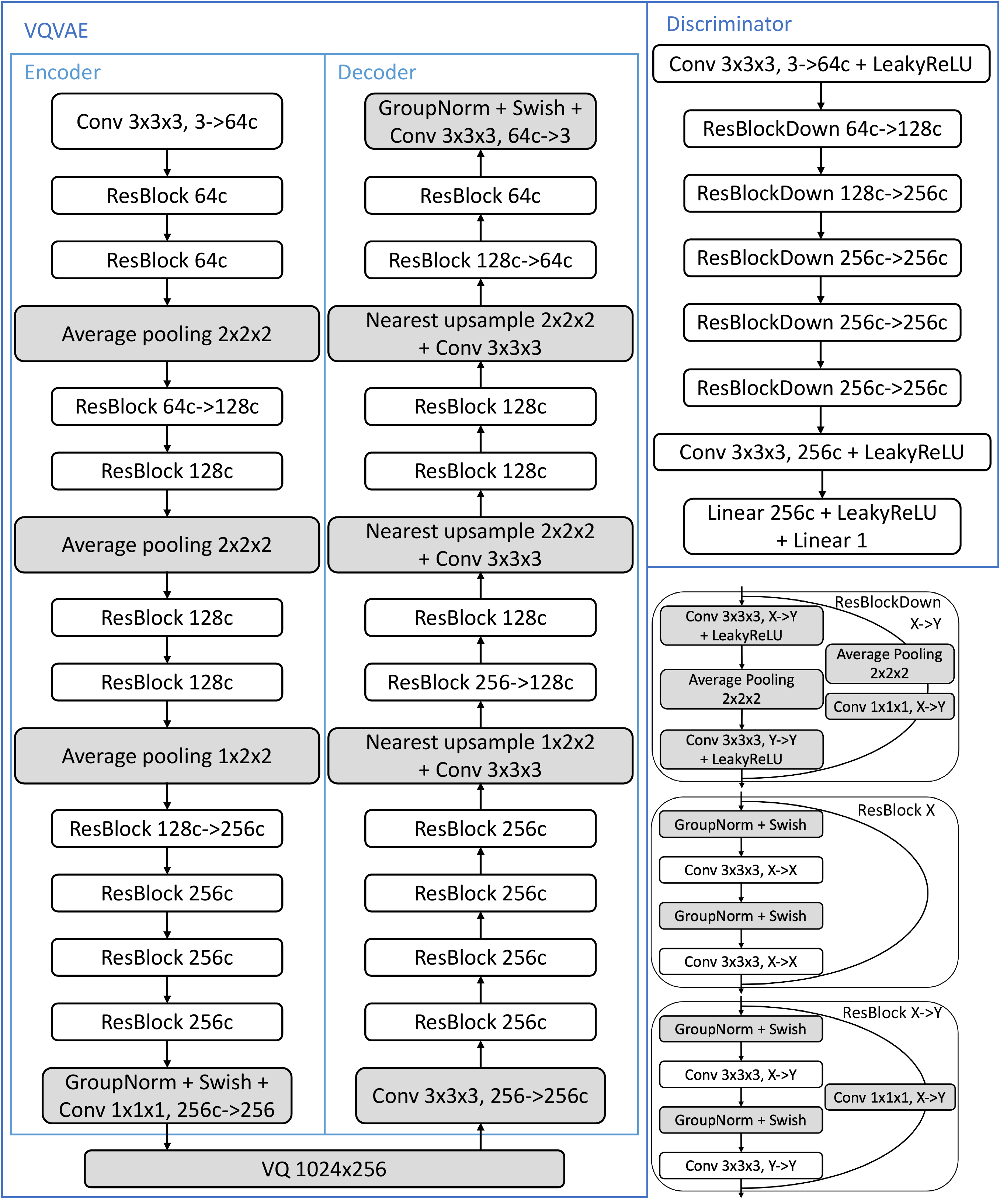}
\caption{\emph{\modelname{}} (ours) 3D-VQ.\\(41M+15M parameters at $c=1$ (B), 158M+61M at $c=2$ (L))}
\end{subfigure}
\caption{\textbf{Comparison of 3D-VQ model architectures between \modelname{} and the TATS~\cite{ge2022long}.} 
We highlight the blocks with major differences in \textcolor{gray}{gray} background and detail their design differences in \cref{app:vq_arch}.
We train the models to quantize 16-frame clips of 128$\times$128 resolution into $4\times16\times16$ tokens.
The number of parameters in parentheses are broken down between VQVAE and discriminators.}
\label{fig:vq_arch}
\end{figure*}

\subsection{3D-VQ Tokenizer}
\label{app:vq_arch}

\cref{fig:vq_arch} shows the architectures of the \modelname{} 3D-VQ module and compares it with the 3D-VQ module in TATS~\cite{ge2022long} which held the previous state-of-the-art for video generation. Compared with TATS, the major design choices in \modelname{} 3D-VQ are listed below. 
\begin{itemize}[nosep]
\item Average pooling, instead of strided convolution, is used for down-sampling.
\item Nearest resizing and convolution
are used for up-sampling.
\item We use spatial down- and up-sampling layers near the latent space and spatial-temporal down- and up-sampling layers near the pixel space, resulting in mirrored encoder-decoder architecture.
\item A single deeper 3D discriminator is designed rather than two shallow discriminators for 2D and 3D separately.
\item We quantize into a much smaller vocabulary of 1,024 as compared to 16,384.
\item We use group normalization~\cite{wu2018group} instead of batch normalization~\cite{ioffe2015batch} and Swish~\cite{ramachandran2018searching} activation function instead of SiLU~\cite{hendrycks2016gaussian}.
\item We use the LeCAM regularization~\cite{tseng2021regularizing} to improve the training stability and quality.
\end{itemize}

The quantitative comparison of the 3D-VQ from TATS and \modelname{} were presented in Table 6 of the main paper. In addition, \cref{fig:ucf_token} below qualitatively compares their reconstruction quality on UCF-101. \cref{fig:vq_sample1,fig:vq_sample2} show \modelname{}'s high-quality reconstruction on example YouTube videos.

We design two variants of the \modelname{} 3D-VQ module, \ie, the base (B) with 41M parameters and the large (L) with 158M parameters, excluding the discriminators.

\subsection{Transformer}
\label{app:bert_arch}
\modelname{} uses the BERT transformer architecture~\cite{devlin2019bert} adapted from the
Flaxformer implementation\footnote{\url{https://github.com/google/flaxformer}}.
Following the transformer configurations in ViT ~\cite{dosovitskiy2020image}, we use two variants of transformers, \ie, base (B) with 87M parameters and large (L) with 306M in all our experiments. \cref{tab:trans_arch} lists the detailed configurations for each variant.
A huge (H) transformer is only used to train on the large Web video dataset and generate demo videos.

\begin{table*}[tp]
\centering
\begin{tabular}{ccccccc}
\toprule
Model     & Param. & \# heads & \# layers & Hidden size & MLP dim \\
\midrule
\modelname-\textbf{B} & 87 M          & 12     & 12     & 768         & 3072    \\
\modelname-\textbf{L} & 305 M         & 16     & 24     & 1024        & 4096    \\
\modelname-\textbf{H} & 634 M         & 16     & 32     & 1280        & 5120    \\
\bottomrule
\end{tabular}
\caption{\textbf{Transformer architecture configurations used in \modelname{}.}}
\label{tab:trans_arch}
\end{table*}


\section{Implementation Details}
\label{app:imp_details}

\subsection{Task Definitions}
\label{app:task}
We employ a total of ten tasks for multi-task video generation. Each task is characterized by a few adjustable settings such as interior condition shape, padding function, and optionally prefix condition. \cref{fig:task} illustrates the interior condition regions for each task under the above setup.
Given a video of shape $T\times H \times W$, we define the tasks as following:

\begin{figure}[tp]
\centering
\begin{subfigure}[t]{0.24\linewidth}
\includegraphics[width=\linewidth]{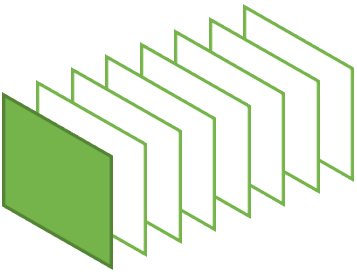}
\caption{FP}
\end{subfigure}
\begin{subfigure}[t]{0.24\linewidth}
\includegraphics[width=\linewidth]{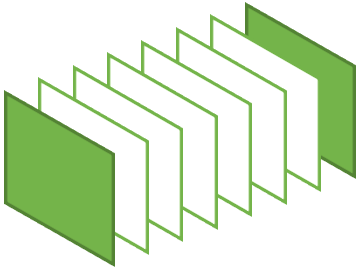}
\caption{FI}
\end{subfigure}
\begin{subfigure}[t]{0.24\linewidth}
\includegraphics[width=\linewidth]{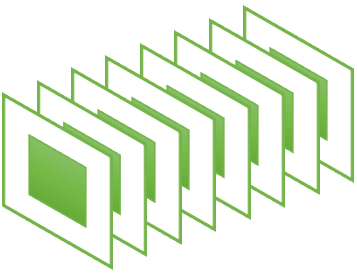}
\caption{OPC}
\end{subfigure}
\begin{subfigure}[t]{0.24\linewidth}
\includegraphics[width=\linewidth]{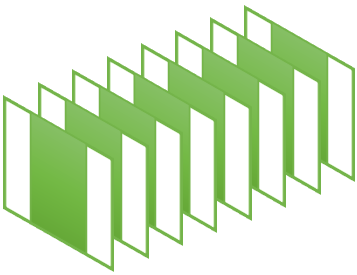}
\caption{OPV}
\end{subfigure}
\begin{subfigure}[t]{0.24\linewidth}
\includegraphics[width=\linewidth]{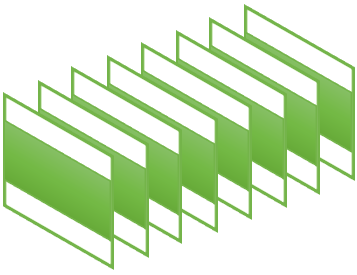}
\caption{OPH}
\end{subfigure}
\begin{subfigure}[t]{0.24\linewidth}
\includegraphics[width=\linewidth]{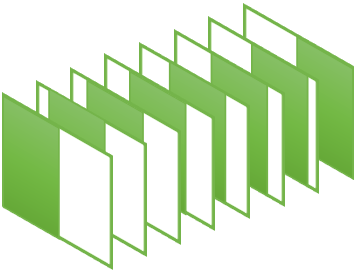}
\caption{OPD}
\end{subfigure}
\begin{subfigure}[t]{0.24\linewidth}
\includegraphics[width=\linewidth]{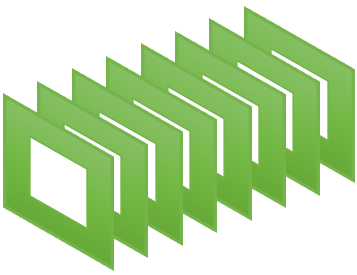}
\caption{IPC}
\end{subfigure}
\begin{subfigure}[t]{0.24\linewidth}
\includegraphics[width=\linewidth]{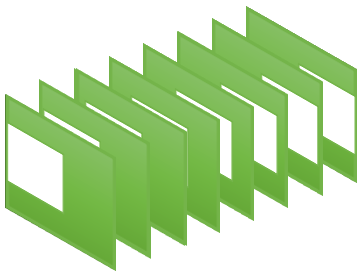}
\caption{IPD}
\end{subfigure}
\begin{subfigure}[t]{0.24\linewidth}
\includegraphics[width=\linewidth]{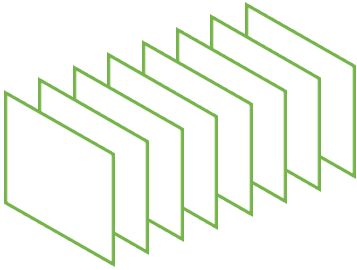}
\caption{CG}
\end{subfigure}
\begin{subfigure}[t]{0.24\linewidth}
\includegraphics[width=\linewidth]{fp_vid}
\caption{CFP}
\end{subfigure}
\caption{\textbf{Interior condition regions for each task}, where \textcolor{ForestGreen}{green} denotes valid pixels and white pixels denote the task-specific paddings discussed in \cref{app:task}. The tasks are Frame Prediction (FP), Frame Interpolation (FI), Central Outpainting (OPC), Vertical Outpainting (OPV), Horizontal Outpainting (OPH), Dynamic Outpainting (OPD), Central Inpainting (IPC), Dynamic Inpainting (IPD), Class-conditional Generation (CG), and Class-conditional Frame Prediction (CFP).}
\label{fig:task}
\end{figure}

\begin{itemize}[nosep, leftmargin=*]
\item Frame Prediction (FP)
\begin{itemize}[nosep, leftmargin=*]
    \item Interior condition: $t$ frames at the beginning; $t=1$.
    \item Padding: replicate the last given frame.
\end{itemize}
\item Frame Interpolation (FI)
\begin{itemize}[nosep, leftmargin=*]
    \item Interior condition: $t_1$ frames at the beginning and $t_2$ frames at the end; $t_1=1$, $t_2=1$.
    \item Padding: linear interpolate between the last given frame at the beginning and the first given frame at the end.
\end{itemize}
\item Central Outpainting (OPC)
\begin{itemize}[nosep, leftmargin=*]
    \item Interior condition: a rectangle at the center with height $h$ and width $w$; $h=0.5H$, $w=0.5W$.
    \item Padding: pad the nearest pixel for each location (\texttt{edge} padding).
\end{itemize}
\item Vertical Outpainting (OPV)
\begin{itemize}[nosep, leftmargin=*]
    \item Interior condition: a centered vertical strip with width $w$; $w=0.5W$.
    \item Padding: \texttt{edge} padding.
\end{itemize}
\item Horizontal Outpainting (OPH)
\begin{itemize}[nosep, leftmargin=*]
    \item Interior condition: a centered horizontal strip with height $h$; $h=0.5H$.
    \item Padding: \texttt{edge} padding.
\end{itemize}
\item Dynamic Outpainting (OPD)
\begin{itemize}[nosep, leftmargin=*]
    \item Interior condition: a moving vertical strip with width $w$; $w=0.5W$.
    \item Direction of movement: left to right.
    \item Padding: zero padding.
\end{itemize}
\item Central Inpainting (IPC)
\begin{itemize}[nosep, leftmargin=*]
    \item Interior condition: everything but a rectangle at the center with height $h$ and width $w$; $h=0.5H$, $w=0.5W$.
    \item Padding: zero padding.
\end{itemize}
\item Dynamic Inpainting (IPD)
\begin{itemize}[nosep, leftmargin=*]
    \item Interior condition: everything but a vertically centered moving rectangle with height $h$ and width $w$; $h=0.5H$, $w=0.5W$.
    \item Direction of movement: left to right.
    \item Padding: zero padding.
\end{itemize}
\item Class-conditional Generation (CG)
\begin{itemize}[nosep, leftmargin=*]
    \item Prefix condition: class label.
\end{itemize}
\item Class-conditional Frame Prediction (CFP)
\begin{itemize}[nosep, leftmargin=*]
    \item Prefix condition: class label.
    \item Interior condition:  $t$ frames at the beginning; $t=1$.
    \item Padding: replicate the last given frame.
\end{itemize}
\end{itemize}




\subsection{Training}
\label{app:train}
\modelname{} is trained in two stages where we first train the 3D-VQ tokenizer and then train the transformer with a frozen tokenizer.
We follow the same learning recipe across all datasets, with the only variation in the number of training epochs.
Here are the training details for both stages:

\begin{itemize}[nosep, leftmargin=*]
    \item 3D-VQ:
    \begin{itemize}[nosep, leftmargin=*]
    \item Video: 16 frames, frame stride 1, 128$\times$128 resolution. \\ (64$\times$64 resolution for BAIR)
    \item Base channels: 64 for B, 128 for L.
    \item VQVAE channel multipliers: 1, 2, 2, 4. \\ (1, 2, 4 for 64$\times$64 resolution).
    \item Discriminator channel multipliers: 2, 4, 4, 4, 4. \\ (2, 4, 4, 4 for 64$\times$64 resolution)
    \item Latent shape: 4$\times$16$\times$16.
    \item Vocabulary size: 1,024.
    \item Embedding dimension: 256.
    \item Initialization: central inflation from a 2D-VQ trained on ImageNet with this setup.
    \item Peak learning rate: $10^{-4}$.
    \item Learning rate schedule: linear warm up and\\cosine decay.
    \item Optimizer: Adam with $\beta_1=0$ and $\beta_2=0.99$.
    \item Generator loss type: Non-saturating.
    \item Generator adversarial loss weight: $0.1$.
    \item Perceptual loss weight: $0.1$.
    \item Discriminator gradient penalty: r1 with cost $10$.
    \item EMA model decay rate: $0.999$.
    \item Batch size: 128 for B, 256 for L.
    \item Speed: 0.41 steps/sec on 16 TPU-v2 chips for B, \\ 0.56 steps/sec on 32 TPU-v4 chips for L.
    \end{itemize}
    \item Transformer:
    \begin{itemize}[nosep, leftmargin=*]
    \item Sequence length: 1026.
    \item Hidden dropout rate: $0.1$.
    \item Attention dropout rate: $0.1$.
    \item Mask rate schedule: cosine.
    \item Peak learning rate: $10^{-4}$.
    \item Learning rate schedule: linear warm up and\\cosine decay.
    \item Optimizer: Adam with $\beta_1=0.9$ and $\beta_2=0.96$.
    \item Weight decay $0.045$.
    \item Label smoothing: $10^{-4}$.
    \item Max gradient norm: $1$.
    \item Batch size: 256.
    \item Speed: 1.24 steps/sec on 16 TPU-v2 chips for B, \\ 2.70 steps/sec on 32 TPU-v4 chips for L.
\end{itemize}
\end{itemize}

Using more hardware resources can speed up the training.
We train \modelname{} models for each dataset separately.
The training epochs for each dataset are listed in \cref{tab:train_step}.

\begin{table}[tp]
\centering
\begin{tabular}{lcccccc}
\toprule
\multirow{2}{*}{Dataset} & \multicolumn{2}{c}{3D-VQ} & \multicolumn{2}{c}{Transformer} \\
& B & L & B & L\\ \midrule
UCF-101 & 500 & 2000 & 2000 & 2000\\
BAIR & 400 & 800 & 400 & 800 \\ 
BAIR-MT & 400 & 800 & 1200 & 1600 \\
Kinetics-600 & 45 & 180 & 180 & 360 \\
SSv2 & 135 & 400 & 720 & 1440 \\
nuScenes & 1280 & 5120 & 2560 & 10240 \\
Objectron & 1000 & 2000 & 1000 & 2000 \\
Web & 5 & 20 & 10 & 20 \\ \bottomrule
\end{tabular}
\caption{\textbf{Training epochs for each dataset.}}
\label{tab:train_step}
\end{table}

\subsection{Evaluation}
\label{app:eval}

\paragraph{Evaluation metrics.}
The FVD~\cite{unterthiner2018towards} is used as the primary evaluation metric. 
We follow the official implementation\footnote{\url{https://github.com/google-research/google-research/tree/master/frechet_video_distance}} in extracting video features with an I3D model trained on Kinetics-400~\cite{carreira2017quo}.
We report Inception Score (IS)~\cite{saito2020train}\footnote{\url{https://github.com/pfnet-research/tgan2}} on the UCF-101 dataset which is calculated with a C3D~\cite{tran2015learning} model trained on UCF-101. 
We further include image quality metrics: PSNR, SSIM~\cite{wang2004image} and LPIPS~\cite{zhang2018unreasonable} (computed by the VGG features) on the BAIR dataset.

\paragraph{Sampling protocols.}
We follow the sampling protocols from previous works~\cite{ge2022long,clark2019adversarial} when eveluating on the standard benchmarks, \ie UCF-101, BAIR, and Kinetics-600.
We sample 16-frame clips from each dataset without replacement to form the real distribution in FVD and extract condition inputs from them to feed to the model.
We continuously run through all the samples required (\eg, 40,000 for UCF-101) with a single data loader and compute the mean and standard deviation for 4 folds.
When evaluating on other datasets, due to the lack of prior works, we adapt the above protocol based on the dataset size to ensure sample diversity.

For our \modelname{} model, we use the following COMMIT decoding hyperparameters by default: cosine schedule, $12$ steps, temperature $4.5$.
Below are detailed setups for each dataset:

\begin{itemize}[nosep, leftmargin=*]
    \item UCF-101: 
    \begin{itemize}[nosep, leftmargin=*]
        \item Dataset: 9.5K videos for training, 101 classes.
        \item Number of samples: 10,000$\times$4.
        \item Resolution: 128$\times$128.
        \item Real distribution: random clips from the training videos.
    \end{itemize}
    \item BAIR:
    \begin{itemize}[nosep, leftmargin=*]
        \item Dataset: 43K videos for training and 256 videos for evaluation.
        \item Number of samples: 25,600$\times$4.
        \item Resolution: 64$\times$64.
        \item Real distribution: the first 16-frame clip from each evaluation video.
        \item COMMIT decoding: exponential schedule, temperature 400.
    \end{itemize}
    \item Kinetics-600:
    \begin{itemize}[nosep, leftmargin=*]
        \item Dataset: 384K videos for training and 29K videos for evaluation.
        \item Number of samples: 50,000$\times$4.
        \item Generation resolution: 128$\times$128.
        \item Evaluation resolution: 64$\times$64, via central crop and bilinear resize.
        \item Real distribution: 6 sampled clips (2 temporal windows and 3 spatial crops) from each evaluation video.
        \item COMMIT decoding: uniform schedule, temperature 7.5.
    \end{itemize}
    \item SSv2:
    \begin{itemize}[nosep, leftmargin=*]
        \item Dataset: 169K videos for training and 24K videos for evaluation, 174 classes.
        \item Number of samples: 50,000$\times$4.
        \item Resolution: 128$\times$128.
        \item Real distribution for the CG task: random clips from the training videos.
        \item Real distribution for the other tasks: 2 sampled clips (2 temporal windows and central crop) from each evaluation video.
    \end{itemize}
    \item nuScenes:
    \begin{itemize}[nosep, leftmargin=*]
        \item Dataset: 5.4K videos for training and 0.6K videos for evaluation, front camera only, 32 frames per video.
        \item Number of samples: 50,000$\times$4.
        \item Resolution: 128$\times$128.
        \item Real distribution: 48 sampled clips (16 temporal windows and 3 spatial crops) from each evaluation video.
    \end{itemize}
    \item Objectron:
    \begin{itemize}[nosep, leftmargin=*]
        \item Dataset: 14.4K videos for training and 3.6K videos for evaluation.
        \item Number of samples: 50,000$\times$4.
        \item Resolution: 128$\times$128.
        \item Real distribution: 5 sampled clips (5 temporal windows and central crop) from each evaluation video.
    \end{itemize}
    \item Web videos:
    \begin{itemize}[nosep, leftmargin=*]
        \item Dataset: $\sim$12M videos for training and 26K videos for evaluation.
        \item Number of samples: 50,000$\times$4.
        \item Resolution: 128$\times$128.
        \item Real distribution: randomly sampled clips from evaluation videos.
    \end{itemize}
\end{itemize}

\begin{table}[tp]
\centering
\begin{tabular}{@{}l@{\hspace{4pt}}c@{\hspace{4pt}}c@{\hspace{4pt}}c@{\hspace{5pt}}c@{}}
\toprule
Method         & \makecell{Extra\\Video}  & Class      & FVD$\downarrow$          & IS$\uparrow$                \\ \midrule
VGAN\cite{vondrick2016generating} & & \checkmark & -                        & 8.31\mytiny{$\pm$0.09}  \\
TGAN\cite{saito2017temporal}& &            & -                        & 11.85\mytiny{$\pm$0.07} \\
MoCoGAN$^*$\cite{tulyakov2018mocogan}& & \checkmark & -                        & 12.42\mytiny{$\pm$0.07} \\
ProgressiveVGAN\cite{acharya2018towards}&& \checkmark & -                        & 14.56\mytiny{$\pm$0.05} \\
TGAN\cite{saito2017temporal}& & \checkmark & -                        & 15.83\mytiny{$\pm$0.18} \\
RaMViD\cite{hoppe2022diffusion}   & &    &   -  &  21.71\mytiny{$\pm$0.21}  \\
LDVD-GAN\cite{kahembwe2020lower} &&            & -                        & 22.91\mytiny{$\pm$0.19} \\
StyleGAN-V$^{*\#}$\cite{skorokhodov2022stylegan}& &  & - & 23.94\mytiny{$\pm$0.73} \\
VideoGPT\cite{yan2021videogpt}& &            & -                        & 24.69\mytiny{$\pm$0.30} \\
TGANv2\cite{saito2020train}& & \checkmark & 1209\mytiny{$\pm$28} & 28.87\mytiny{$\pm$0.67} \\
MoCoGAN-HD$^\#$\cite{tian2021good}& &            & 838                      & 32.36                       \\
DIGAN\cite{yu2022generating} &&            & 655\mytiny{$\pm$22}  & 29.71\mytiny{$\pm$0.53} \\

DIGAN$^\#$\cite{yu2022generating}& &            & 577\mytiny{$\pm$21}  & 32.70\mytiny{$\pm$0.35} \\
DVD-GAN$^\#$\cite{clark2019adversarial}& & \checkmark & -                        & 32.97\mytiny{$\pm$1.70}  \\
Video Diffusion$^{*\#}$\cite{ho2022video}&&            & -                        & 57.00\mytiny{$\pm$0.62}    \\
TATS\cite{ge2022long} & &            & 420\mytiny{$\pm$18}  & 57.63\mytiny{$\pm$0.24} \\
CCVS+StyleGAN$^\#$\cite{le2021ccvs} &&            & 386\mytiny{$\pm$15}  & 24.47\mytiny{$\pm$0.13} \\
Make-A-Video$^*$\cite{singer2022make} & & \checkmark & 367 & 33.00 \\
TATS\cite{ge2022long} && \checkmark & 332\mytiny{$\pm$18}  & 79.28\mytiny{$\pm$0.38} \\ \midrule
\color{gray}CogVideo$^*$\cite{hong2022cogvideo} & \color{gray}\checkmark  & \color{gray}\checkmark & \color{gray}626                      & \color{gray}50.46                       \\
\color{gray}Make-A-Video$^*$\cite{singer2022make} & \color{gray}\checkmark & \color{gray}\checkmark & \color{gray}81 & \color{gray}82.55 \\ \midrule
\emph{\modelname}-B-CG (ours) & & \checkmark & \underline{159\mytiny{$\pm$2}} & \underline{83.55\mytiny{$\pm$0.14}}     \\
\emph{\modelname}-L-CG (ours) & & \checkmark & \textbf{76\mytiny{$\pm$2}}     & \textbf{89.27\mytiny{$\pm$0.15}}     \\ \bottomrule
\end{tabular}
\caption{\textbf{Generation performance on the UCF-101 dataset.} 
Methods in \textcolor{gray}{gray} are pretrained on additional large video data.
Methods with $\checkmark$ in the Class column are class-conditional, while the others are unconditional. Methods marked with $^*$ use custom resolutions, while the others are at 128$\times$128.
Methods marked with $^\#$ additionally used the test set in training. }
\label{tab:ucf_full}
\end{table}
\begin{table}[tp]
\centering
\begin{tabular}{@{}lcc@{}}
\toprule
Method                   & K600 FVD$\downarrow$   & BAIR FVD$\downarrow$ \\ \midrule
LVT\cite{rakhimov2021latent} & 224.7                      & 126\mytiny{$\pm$3} \\
Video Transformer\cite{weissenborn2019scaling} & 170.0\mytiny{$\pm$5.0} & 94\mytiny{$\pm$2}  \\
CogVideo$^*$\cite{hong2022cogvideo}  & 109.2                      & -                      \\
DVD-GAN-FP\cite{clark2019adversarial} & 69.1\mytiny{$\pm$1.2}  & 110                    \\
CCVS\cite{le2021ccvs}    & 55.0\mytiny{$\pm$1.0}  & 99\mytiny{$\pm$2}  \\
Phenaki\cite{villegas2022phenaki} & 36.4\mytiny{$\pm$0.2} & 97 \\
VideoGPT\cite{yan2021videogpt}  & -                          & 103                    \\
TrIVD-GAN-FP\cite{luc2020transformation} & 25.7\mytiny{$\pm$0.7}  & 103                    \\
Transframer\cite{nash2022transframer}  & 25.4                       & 100                    \\
MaskViT\cite{gupta2022maskvit} & -                          & 94                     \\
FitVid\cite{babaeizadeh2021fitvid} & -                          & 94                     \\
MCVD\cite{voleti2022masked} & - & 90 \\
N\"UWA\cite{wu2021n}    & -                          & 87                     \\
RaMViD\cite{hoppe2022diffusion}        &  16.5     &   84 \\
Video Diffusion\cite{ho2022video}  & \underline{16.2\mytiny{$\pm$0.3}}  & -                     \\ \midrule
\emph{\modelname}-B-FP (ours)          & 24.5\mytiny{$\pm$0.9}     & \underline{76{\mytiny$\pm$0.1}} (47{\mytiny$\pm$0.1})   \\
\emph{\modelname}-L-FP (ours)         & \textbf{9.9\mytiny{$\pm$0.3}}     & \textbf{62{\mytiny$\pm$0.1}}  (31{\mytiny$\pm$0.2})   \\ \bottomrule
\end{tabular}
\caption{\textbf{Frame prediction performance on the BAIR and Kinetics-600 datasets.}
- marks that the value is unavailable in their paper or incomparable to others.
The FVD in parentheses uses a debiased evaluation protocol on BAIR detailed in \cref{app:eval}.
Methods marked with $^*$ is pretrained on additional large video data.
}
\label{tab:bairk600_full}
\end{table}

For the ``random clips" above, we refer to the combination of a random temporal window and a random spatial crop on a random video.
For the fixed number of ``temporal windows" or ``spatial crops", deterministic uniform sampling is used.

For the image quality metrics on BAIR in Table 3 of the main paper, 
CCVS~\cite{le2021ccvs} generates at $256 \times 256$ while the others are at $64 \times 64$.
When calculating PSNR and SSIM, we follow~\cite{voleti2022masked} in using the best value from 100 trials for each evaluation video.

\paragraph{Debiased FVD on BAIR}
Computing FVD is difficult on the BAIR dataset due to its small evaluation target of only 256 16-frame clips. Following the standard evaluation protocol, we generate 100 predictions for each clip to create 256,00 samples~\cite{babaeizadeh2021fitvid}.


The real distribution to compute FVD in this way is highly biased with the insufficient evaluation videos~\cite{unterthiner2018towards}. We can see this by a simple experiment where we compute the training FVD with only 256 training videos. 
We observe that this 256-sample training FVD ($64$) is far worse than the regular training FVD with all 43K videos ($13$), showing the biased FVD computation.

To bridge the gap, we use uniformly sampled 16-frame clips from the 256 30-frame evaluation videos, which results in $256 \times 15 = 3840$ clips.
The uniform sampling yields a better representation of the evaluation set.
Under this new protocol, \modelname{}-L-FP achieves FVD 31 instead of 62, which is more aligned with its training set performance (FVD=8).

We report this ``debiased FVD'' in addition to the standard FVD computation on the BAIR dataset, with the default COMMIT decoding hyperparameters. 
We also use it for BAIR multi-task evaluation and ablation studies on BAIR .


\section{Additional Quantitative Evaluation}
\label{app:quant}
\begin{table*}[tp]
\centering
\begin{tabular}{@{}ccccccccccccc@{}}
\toprule
Method & Task & \textbf{Avg}$\downarrow$ & FP & FI & OPC & OPV & OPH & OPD  & IPC  & IPD & CG & CFP    \\ \midrule
\modelname-B-UNC & Single & 258.8   & {\color{gray}278.8}  & {\color{gray}91.0} & {\color{gray}67.5} & {\color{gray}27.3}  & {\color{gray}36.2}& {\color{gray}711.5} & {\color{gray}319.3}   & {\color{gray}669.8}  & {\color{gray}107.7}& {\color{gray}279.0} \\
\modelname-B-FP & Single & 402.9   & 59.3 & {\color{gray}76.2} & {\color{gray}213.2} & {\color{gray}81.2} & {\color{gray}86.3}& {\color{gray}632.7}    & {\color{gray}343.1}  & {\color{gray}697.9}  & {\color{gray}1780.0}& 59.3  \\ \midrule
\modelname-B-\emph{MT} & Multi & 43.4 & 71.5 & 38.0 & 38.8 & 23.3 & 26.1 & 33.4  & 23.3  & 25.3 & 94.7 & 59.3   \\
\modelname-L-\emph{MT} & Multi & \textbf{27.3} & 33.8 & 25.0 & 21.1 & 16.8 & 17.0& 23.5  & 13.5  & 15.0  & 79.1 & 28.5   \\ \midrule
Masked pixel & - & - & 94\% & 87\% & 75\% & 50\% & 50\% & 50\% & 25\% & 25\% & 100\% & 94\% \\
Masked token & - & - & 75\% & 50\% & 75\% & 50\% & 50\% & 50\% & 25\% & 25\% & 100\% & 75\% \\
\bottomrule
\end{tabular}
\caption{\textbf{Multi-task generation performance on Something-Something-V2 evaluated by FVD.} \textcolor{gray}{Gray} values denote unseen tasks during training. The bottom two rows list the proportions of masked pixels and tokens for each task.}
\label{tab:ssv2}
\end{table*}

\begin{table*}[tp]
\centering
\begin{tabular}{@{}c|c|c|ccccccccccc@{}}
\toprule
Method & \makecell{nuScenes-FP} & \makecell{Objectron-FI} & \makecell{Web-\textbf{MT8}} & FP & FI & OPC & OPV & OPH & OPD  & IPC & IPD \\ \midrule
\modelname-B & 29.3 & - & 33.0 & 84.9 & 33.9 & 34.4 & 21.5 & 22.1& 26.0  & 20.7  & 20.4    \\
\modelname-L & 20.6 & 26.7 & \textbf{21.6} & 45.5 & 30.9 & 19.9 & 15.3 & 14.5 & 20.2  & 12.0  & 14.7    \\
\bottomrule
\end{tabular}
\caption{\textbf{Generation performance on NuScenes, Objectron, and Web videos evaluated by FVD.}}
\label{tab:other_data}
\end{table*}

\paragraph{Class-conditional generation.}
\cref{tab:ucf_full} shows a detailed comparison with the previously published results on the UCF-101~\cite{soomro2012ucf101} class-conditional video generation benchmark, where the numbers are quoted from the cited papers. 
Note that CogVideo~\cite{hong2022cogvideo} and Make-A-Video~\cite{singer2022make} are pretrained on additional 5-10M videos before finetuning on UCF-101, where Make-A-Video further uses a text-image prior trained on a billion text-image pairs.
The remaining models, including \modelname{}, are only trained on 9.5K training videos of UCF-101, or 13.3K training and testing videos of UCF-101 for those marked with $^\#$. 
\cref{fig:ucf_sample} provides a visual comparison to the baseline methods.

As shown, even the smaller \modelname{}-B performs favorably against previous state-of-the-art model TATS~\cite{ge2022long} by a large margin.
\modelname{}-L pushes both the FVD ($332 \rightarrow 76$, $\downarrow 77\%$) and IS ($79.28 \rightarrow 89.27$, $\uparrow 13\%$) to a new level, while outperforming the contemporary work Make-A-Video~\cite{singer2022make} which is pretrained on significantly large extra training data.

\begin{table*}[tp]
\centering
\begin{tabular}{lccccccccc}
\toprule
\multirow{2}{*}{VQ Tokenizer} & \multicolumn{3}{c}{From Scratch} & \multicolumn{6}{c}{ImageNet Initialization} \\
& PSNR$\uparrow$ & SSIM$\uparrow$ & LPIPS$\downarrow$ & PSNR$\uparrow$ & SSIM$\uparrow$  & LPIPS$\downarrow$ & PSNR$\uparrow$ & SSIM$\uparrow$ & LPIPS$\downarrow$ \\ \midrule
MaskGIT 2D  &
 21.4 & 0.667 & 0.139 & 21.5 & 0.685 & 0.114 & \multicolumn{3}{c}{-} \\
\midrule
& & & & \multicolumn{3}{c}{Average} & \multicolumn{3}{c}{Central} \\
\emph{\modelname{}} 3D-L &
21.8 & 0.690 & 0.113 & 21.9 & 0.697 & 0.103 & \textbf{22.0} & \textbf{0.701} & \textbf{0.099} \\
\bottomrule
\end{tabular}
\caption{\textbf{Image quality metrics of different tokenizers} on UCF-101 training set reconstruction.
}
\label{tab:token_imquality}
\end{table*}

\paragraph{Frame prediction.}
For the frame prediction task on BAIR Robot Pushing~\cite{ebert2017self,unterthiner2018towards} (1-frame condition) and Kinetics-600~\cite{carreira2018short} (5-frame condition), \cref{tab:bairk600_full} provides a detailed comparison with previously published results. We use ``-'' to mark the FVDs that either is unavailable in their paper or incomparable to others. For example, Video Diffusion~\cite{ho2022video}'s FVD reported in their paper was on a different camera angle (top-down view \texttt{image\_main}\footnote{\url{https://www.tensorflow.org/datasets/catalog/bair\_robot\_pushing\_small}}) and is hence incomparable to others.

\modelname{} achieves state-of-the-art quality in terms of FVD on both datasets, with a $39\%$ relative improvement on the large-scale Kinetics benchmark than the highly-competitive Video Diffusion baseline~\cite{ho2022video}. \cref{fig:bair} and \cref{fig:k600} below provide visual comparisons to the baseline methods on BAIR and Kinetics-600, respectively.


\paragraph{Multi-task video generation.}
Having verified single-task video generation, \cref{tab:ssv2} shows per-task performance of the ten tasks on the large-scale Something-Something-v2 (SSv2)~\cite{goyal2017something} dataset, with the proportions of masks in both pixel and token spaces. 
SSv2 is a challenging dataset commonly used for action recognition, whereas this work benchmarks video generation on it for the first time. On this dataset, a model needs to synthesize 174 basic actions with everyday objects. \cref{fig:mt_result_1} shows examples of generated videos for each task on this dataset.

We compare the multi-task models (MT) with two single-task baselines trained on unconditional generation (UNC) and frame prediction (FP).
The multi-task models show consistently better average FVD across all tasks compared with the single-task baselines.

\paragraph{Results on nuScenes, Objectron, and 12M Web Videos.}
\cref{tab:other_data} shows the generation performance on three additional datasets, \ie, nuScenes~\cite{caesar2020nuscenes}, Objectron~\cite{ahmadyan2021objectron}, and 12M Web videos which contains 12 million videos we collected from the web.
We evaluate our model on the frame prediction task on nuScenes, the frame interpolation task on Objectron, and the 8-task suite on the Web videos. \cref{fig:mt_result_2} shows examples of generated videos for each task. The results substantiate the generalization performance of \modelname{} on videos from distinct visual domains and the multi-task learning recipe on large-scale data.

\paragraph{Tokenizer reconstruction.}
We report the image quality metrics (PSNR, SSIM, LPIPS) for the VQGAN reconstruction in  \cref{tab:token_imquality}. 
We compare MAGVIT 3D against the baseline MaskGIT 2D to highlight our 3D design while keeping the remaining components the same. 
As shown, the results in \cref{tab:token_imquality} are consistent with the findings by FVD in Tab. 6.



\section{Qualitative Examples}
\label{app:qualitative}


\subsection{High-Fidelity Tokenization}

\paragraph{Comparison of tokenizers.}

\begin{figure*}[p]
\centering
\begin{subfigure}[t]{\linewidth}
\centering
\includegraphics[width=0.7\linewidth]{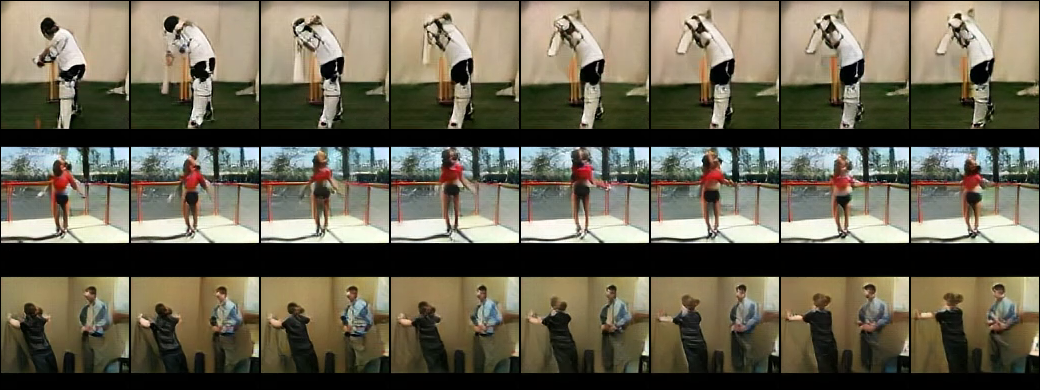}
\caption{MaskGIT~\cite{chang2022maskgit} 2D-VQ
}
\end{subfigure}
\hfill
\begin{subfigure}[t]{\linewidth}
\centering
\includegraphics[width=0.7\linewidth]{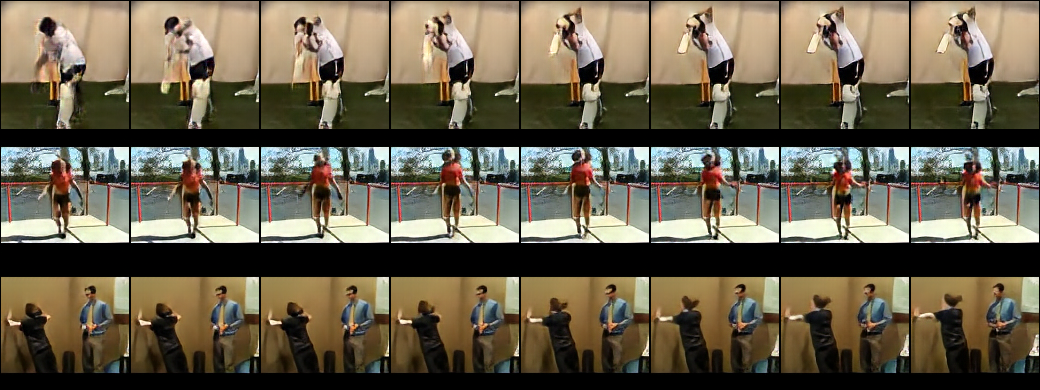}
\caption{TATS~\cite{ge2022long} 3D-VQ
}
\end{subfigure}
\hfill
\begin{subfigure}[t]{\linewidth}
\centering
\includegraphics[width=0.7\linewidth]{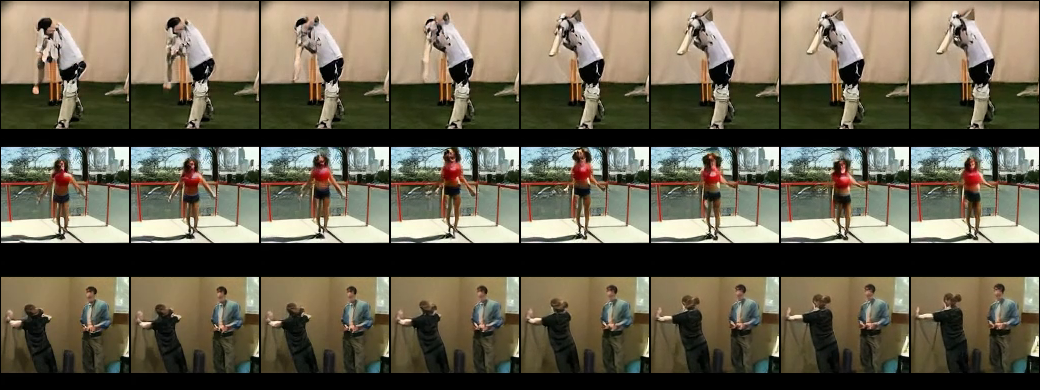}
\caption{\modelname{} 3D-VQ-L (ours)
}
\end{subfigure}
\hfill
\begin{subfigure}[t]{\linewidth}
\centering
\includegraphics[width=0.7\linewidth]{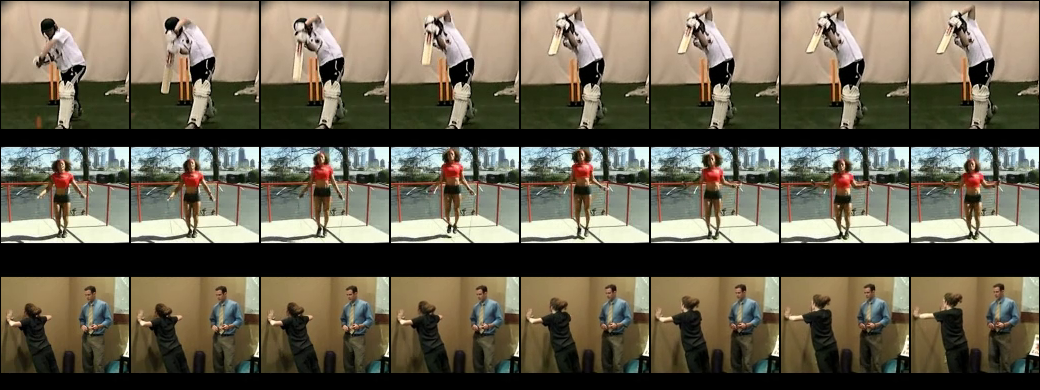}
\caption{Real
}
\end{subfigure}
\caption{\textbf{Comparison of tokenizers on UCF-101 training set reconstruction.}
Videos are reconstructed at 16 frames 64$\times$64 resolution 25 fps and shown at 12.5 fps, with the ground truth in (d). MaskGIT 2D-VQ produces a reasonable image quality, but falls short of frame consistency which causes significant flickering when played as a video (\eg, the curtain color in the first row and the wall color in the third row).
TATS 3D-VQ has a better temporal consistency but loses details for moving objects (\eg, the woman's belly in the second row).
In contrast, our 3D VQ produces consistent frames with greater details reconstructed for both static and moving pixels.
}
\label{fig:ucf_token}
\end{figure*}

\cref{fig:ucf_token} compares the reconstruction quality of the three VQ tokenizers on the UCF-101, including the 2D-VQ from MaskGIT~\cite{chang2022maskgit}, the 3D-VQ from TATS~\cite{ge2022long}, and \modelname{} 3D-VQ, where the videos are taken from the UCF-101 training set. We obtain the TATS model from their official release~\footnote{\url{https://songweige.github.io/projects/tats/}}. We train the MaskGIT 2D-VQ and \modelname{} 3D-VQ using the same protocol on the UCF-101 dataset.

We can see that the MaskGIT 2D-VQ produces a reasonable image quality, but falls short of frame consistency which causes significant flickering when played as a video (\eg, the curtain color in the first row and the wall color in the third row).
TATS 3D-VQ has a better temporal consistency but loses details for moving objects (\eg, the woman's belly in the second row).
In contrast, our 3D VQ produces consistent frames with greater details reconstructed for both static and moving pixels.

\paragraph{Scalable tokenization.}
Since the tokenizers are trained in an unsupervised manner, they exhibit remarkable generalization performances and can be scaled to big data as no labels are required. To demonstrate this, we train a large \modelname{} 3D-VQ on the large YouTube-8M~\cite{abu2016youtube} dataset while ignoring the labels, and use the model to quantize randomly sampled videos on YouTube.

\begin{figure*}[p]
\centering
\includegraphics[height=0.87\linewidth,trim={0 13.7cm 0 0},clip,angle=-90]{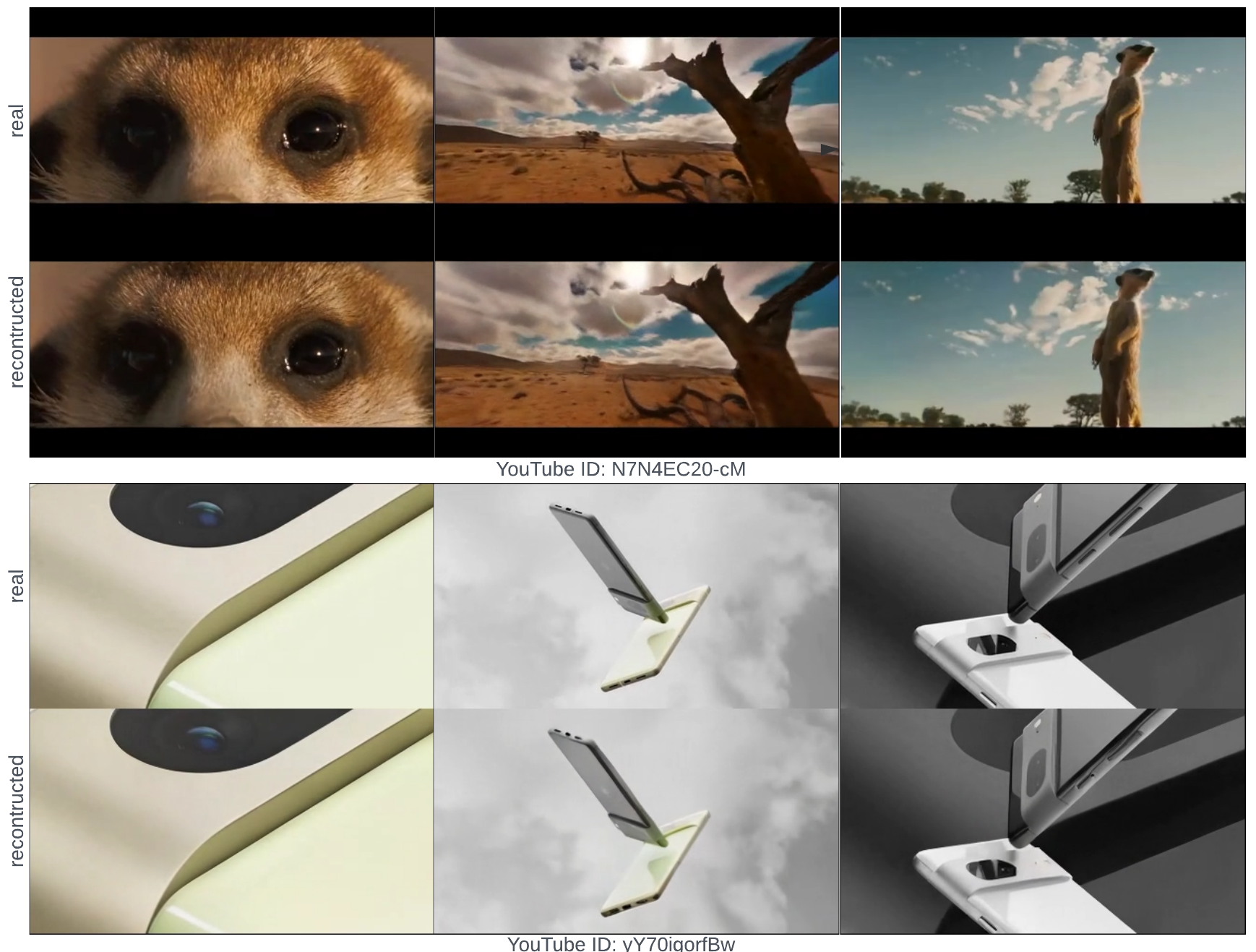}
\caption{\textbf{Our 3D-VQ model produces high reconstruction fidelity with scalable spatial-temporal resolution.} For each group, the top row contains real YouTube videos and the bottom row shows the reconstructed videos from the discrete tokens. 
The original videos are in 240p (240 $\times$ 432) resolution with $N$ frames.
Our 3D-VQ model represents the video as $\frac{N}{4}$ $\times$30$\times$ 54 discrete tokens with a codebook of size 1024, representing a total compression rate of 614.4.
Despite such high compression, the reconstructed results show stunning details and are almost indistinguishable from the real videos.}
\label{fig:vq_sample1}
\end{figure*}

\begin{figure*}[p]
\centering
\includegraphics[height=0.87\linewidth,trim={0 0 0 13.7cm},clip,angle=-90]{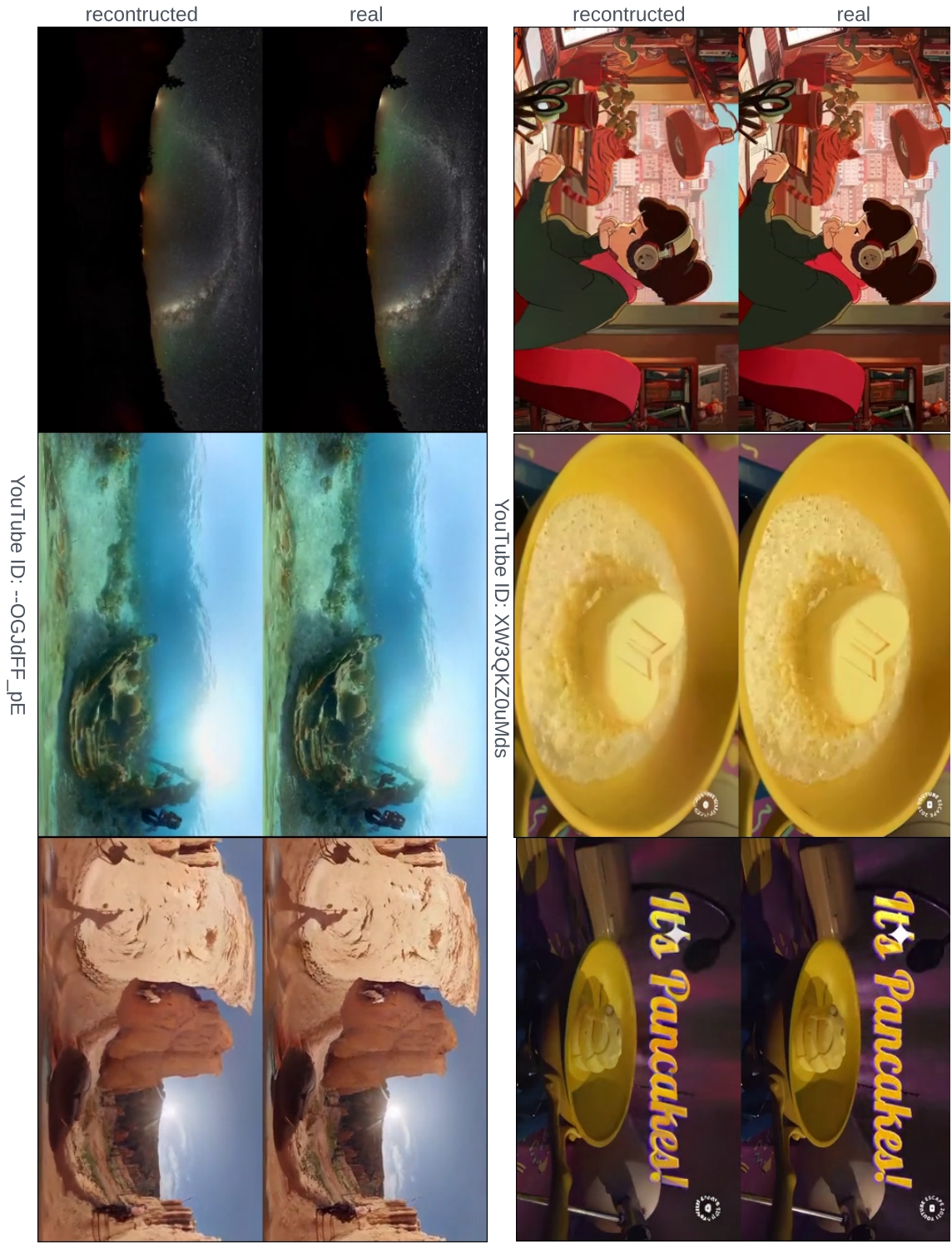}
\caption{\textbf{Our 3D-VQ model produces high reconstruction fidelity with scalable spatial-temporal resolution.} For each group, the top row contains real YouTube videos and the bottom row shows the reconstructed videos from the discrete tokens. 
The original videos are in 240p (240 $\times$ 432) resolution with $N$ frames.
Our 3D-VQ model represents the video as $\frac{N}{4}$ $\times$30$\times$ 54 discrete tokens with a codebook of size 1024, representing a total compression rate of 614.4.
Despite such high compression, the reconstructed results show stunning details and are almost indistinguishable from the real videos.}
\label{fig:vq_sample2}
\end{figure*}

\cref{fig:vq_sample1,fig:vq_sample2} show the original and reconstructed videos from YouTube at 240p (240 $\times$ 432) resolution with arbitrary lengths (\eg 4,096 frames).
Although the tokenizer is only trained with 16-frame 128$\times$128 videos, it produces high reconstruction fidelity for high spatial-temporal resolutions that are unseen in training. Our 3D-VQ model compresses the video by a factor of 4 temporally, by 8$\times$8 spatially, and by 2.4 (24 bits $\rightarrow$ 10 bits) per element, yielding a 614.4$\times$ compression rate.
Despite such high compression, the reconstructed results show stunning details and are almost indistinguishable from the real videos.


\subsection{Single-Task Generation Examples}

\begin{figure*}[p]
\centering
\begin{subfigure}[t]{\linewidth}
\centering
\includegraphics[width=0.7\linewidth]{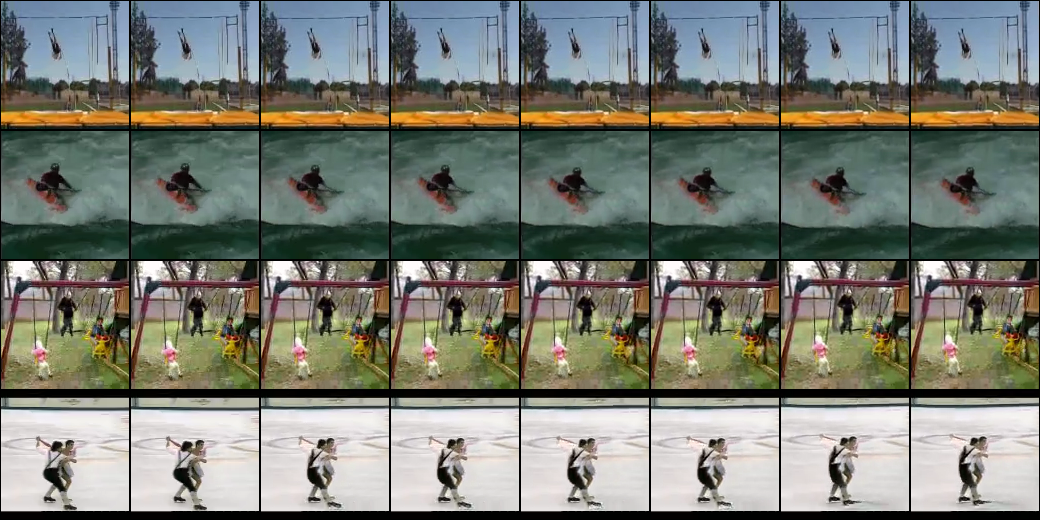}
\caption{CCVS+StyleGAN~\cite{le2021ccvs}
}
\end{subfigure}
\hfill
\begin{subfigure}[t]{\linewidth}
\centering
\includegraphics[width=0.7\linewidth]{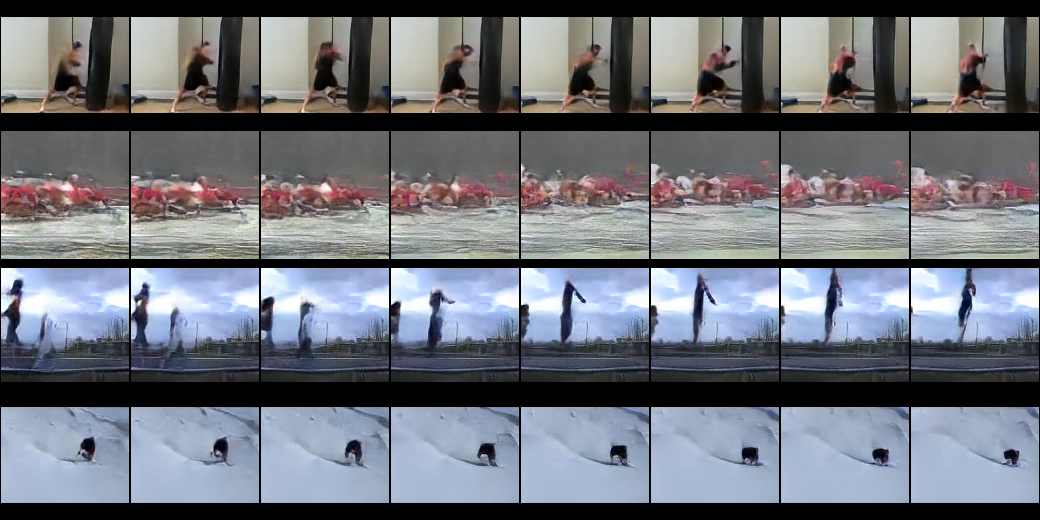}
\caption{TATS~\cite{ge2022long}
}
\end{subfigure}
\hfill
\begin{subfigure}[t]{\linewidth}
\centering
\includegraphics[width=0.7\linewidth]{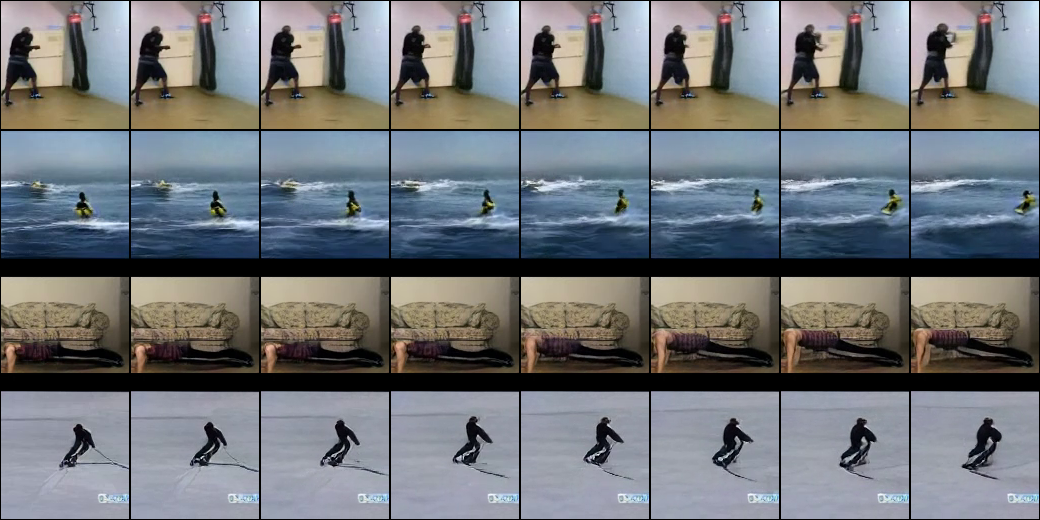}
\caption{\modelname{}-L-CG (ours)
}
\end{subfigure}
\caption{\textbf{Comparison of class-conditional generation samples on UCF-101.} 16-frame videos are generated at 128$\times$128 resolution 25 fps and shown at 12.5 fps. Samples for \cite{le2021ccvs,ge2022long} are obtained from their official release (\url{https://songweige.github.io/projects/tats/}).
CCVS+StyleGAN gets a decent single-frame quality attributing to the pretrained StyleGAN, 
but yields little or no motion.
TATS generates some motion but with clear artifacts.
In contrast, our model produces higher-quality frames with substantial motion.
}
\label{fig:ucf_sample}
\end{figure*}

\cref{fig:ucf_sample} compares the generated samples from CCVS+StyleGAN~\cite{le2021ccvs}, the prior state-of-the-art TATS~\cite{ge2022long}, and \modelname{} on the UCF-101 class-conditional generation benchmark.
As shown in \cref{fig:ucf_sample}, CCVS+StyleGAN~\cite{le2021ccvs} gets a decent single-frame quality attributing to the pretrained StyleGAN, 
but yields little or no motion.
TATS~\cite{ge2022long} generates some motion but with clear artifacts.
In contrast, our model produces higher-quality frames with substantial motion.

\begin{figure*}[tp]
\centering
\begin{subfigure}[t]{0.7\linewidth}
\includegraphics[width=\linewidth]{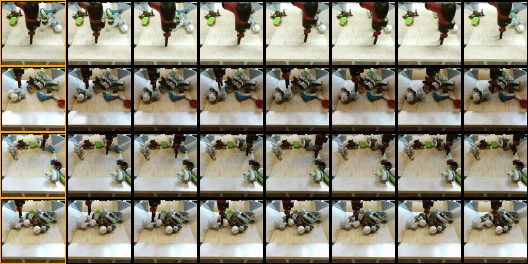}
\caption{RaMViD~\cite{hoppe2022diffusion}}
\end{subfigure}
\begin{subfigure}[t]{0.7\linewidth}
\includegraphics[width=\linewidth]{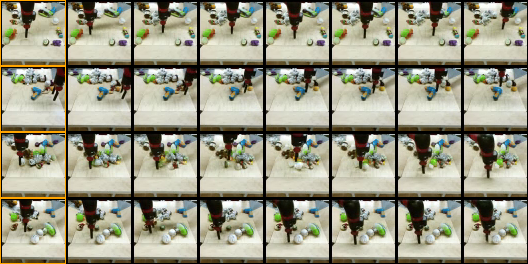}
\caption{\modelname-L-FP (ours)}
\end{subfigure}
\caption{\textbf{Comparison of frame prediction samples on BAIR unseen evaluation set.} 16-frame videos are generated at 64$\times$64 resolution 10 fps given the first frame as condition and shown at 5 fps where condition frames are marked in \textcolor{orange}{orange}. Samples for \cite{hoppe2022diffusion} are obtained from their official release (\url{https://sites.google.com/view/video-diffusion-prediction}).
As shown, the clips produced by \modelname{} maintaining a better visual consistency and spatial-temporal dynamics.}
\label{fig:bair}
\end{figure*}

\cref{fig:bair} compares the generated samples between the state-of-the-art RaMViD~\cite{hoppe2022diffusion} and \modelname{} on the BAIR frame prediction benchmark given 1-frame condition.  As shown, the clips produced by \modelname{} maintaining a better visual consistency and spatial-temporal dynamics.

\begin{figure*}[tp]
\centering
\begin{subfigure}[t]{0.83\linewidth}
\centering
\includegraphics[width=\linewidth]{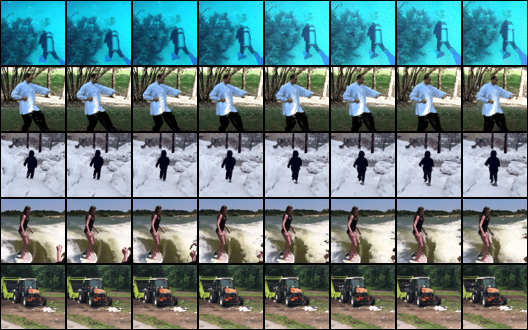}
\caption{RaMViD~\cite{hoppe2022diffusion} at 64$\times$64 resolution, condition information is unavailable.}
\end{subfigure}
\begin{subfigure}[t]{0.83\linewidth}
\centering
\includegraphics[width=\linewidth]{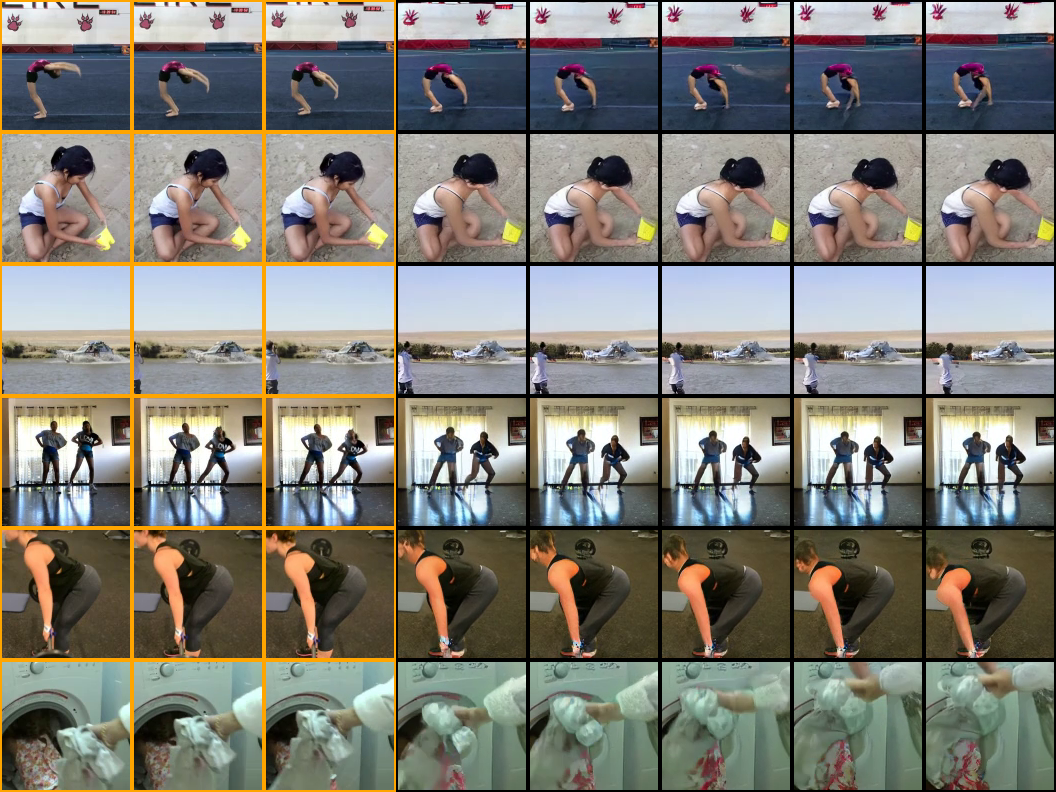}
\caption{\modelname-L-FP (ours) at 128$\times$128 resolution, condition frames are marked in \textcolor{orange}{orange}.}
\end{subfigure}
\caption{\textbf{Comparison of frame prediction samples on Kinetics-600 unseen evaluation set.} 16-frame videos are generated at 25 fps given 5-frame condition. Samples for \cite{hoppe2022diffusion} are obtained from their official release (\url{https://sites.google.com/view/video-diffusion-prediction}).
As shown, given the conditioned frames, \modelname{} generates plausible actions with greater details.}
\label{fig:k600}
\end{figure*}

\cref{fig:k600} compares the generated samples from RaMViD~\cite{hoppe2022diffusion} and \modelname{} on the Kinetics-600 frame prediction benchmark given 5-frame condition. Note that RaMViD generates video in 64$\times$64 and \modelname{} in 128$\times$128 where the standard evaluation is carried out on 64$\times$64. As shown, given the conditioned frames, \modelname{} generates plausible actions with greater details.

\subsection{Multi-Task Generation Examples}

\begin{figure*}[p]
\centering
\includegraphics[width=\linewidth,trim={0 0 0 0},clip]{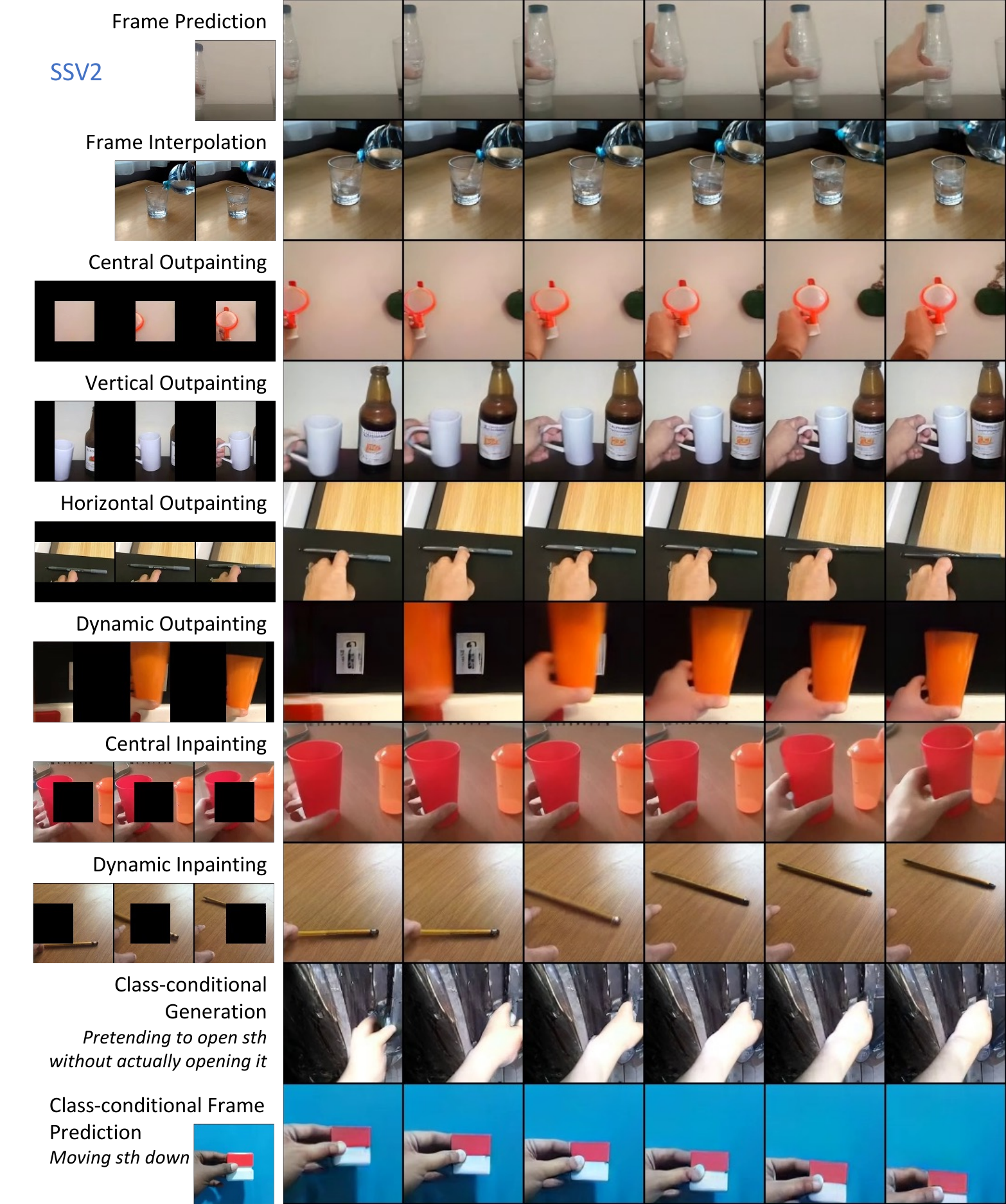}
\caption{\textbf{Multi-task generation results} for the model only trained on the Something-Something-V2 dataset~\cite{goyal2017something}. The condition used to generate the shown videos are taken from the Something-Something-V2 evaluation videos.}
\label{fig:mt_result_1}
\end{figure*}

\begin{figure*}[p]
\centering
\includegraphics[width=\linewidth,trim={0 0 0 0},clip]{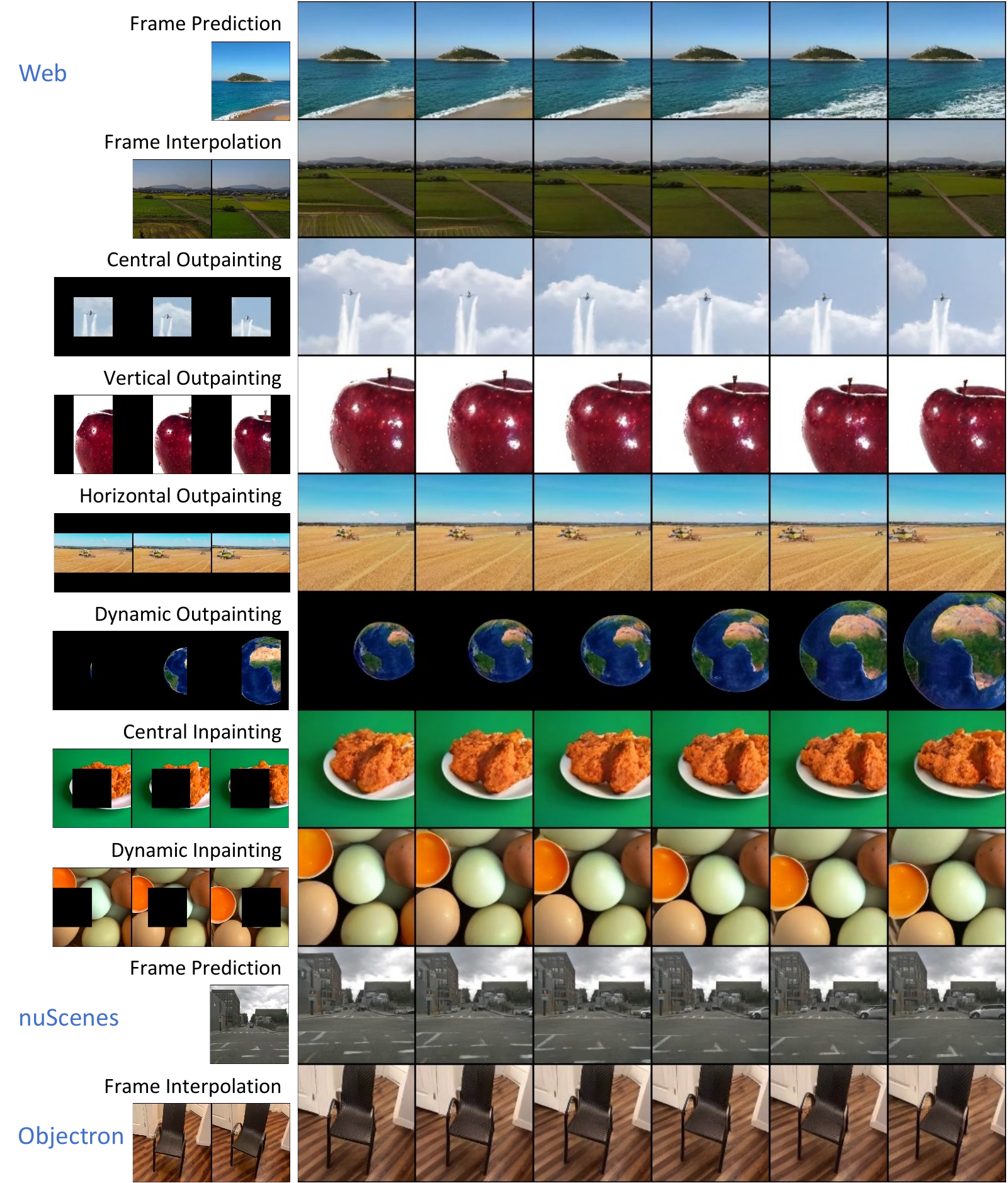}
\caption{\textbf{Multi-task generation results} for three models trained on nuScenes~\cite{caesar2020nuscenes}, Objectron~\cite{ahmadyan2021objectron}, and 12M Web videos, respectively. The condition used to generate the shown videos are taken from the evaluation set.}
\label{fig:mt_result_2}
\end{figure*}

\cref{fig:mt_result_1} shows multi-task generation results on 10 different tasks from a single model trained on SSv2.
\cref{fig:mt_result_2} shows multi-task samples for three other models trained on nuScenes, Objectron, and Web videos.
These results substantiate the multi-task flexibility of \modelname{}.

\begin{figure*}[p]
\centering
\includegraphics[width=\linewidth,trim={0 27.7cm 0 0},clip]{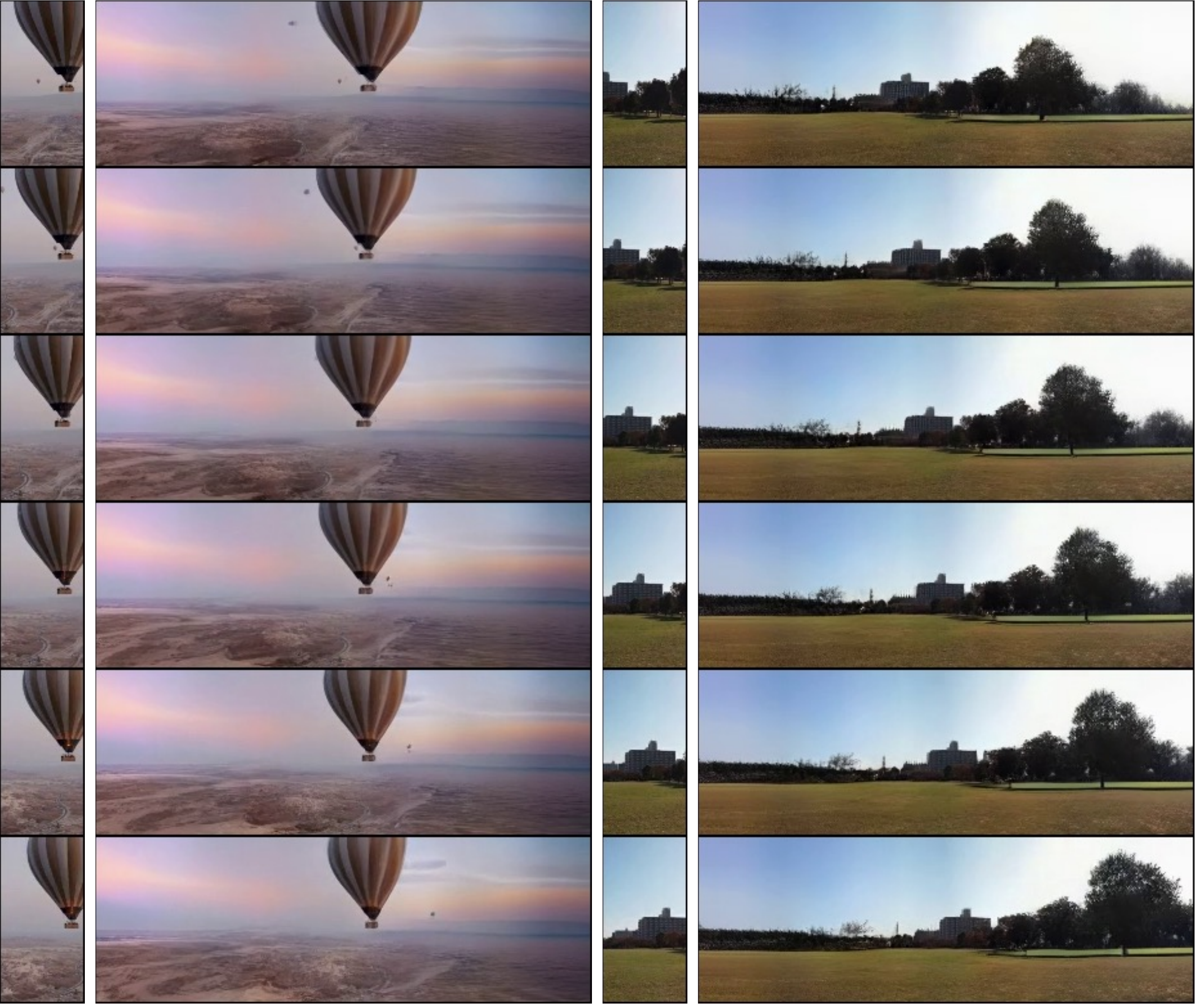}
\caption{\textbf{Ultrawide outpainting results}. Given a vertical slice of 64$\times$128, \modelname{} expands it into a panorama video of 384$\times$128 by doing vertical outpainting for 5 times on each side.}
\label{fig:wide_op1}
\end{figure*}

\begin{figure*}[p]
\centering
\includegraphics[width=\linewidth,trim={0 0 0 27.7cm},clip]{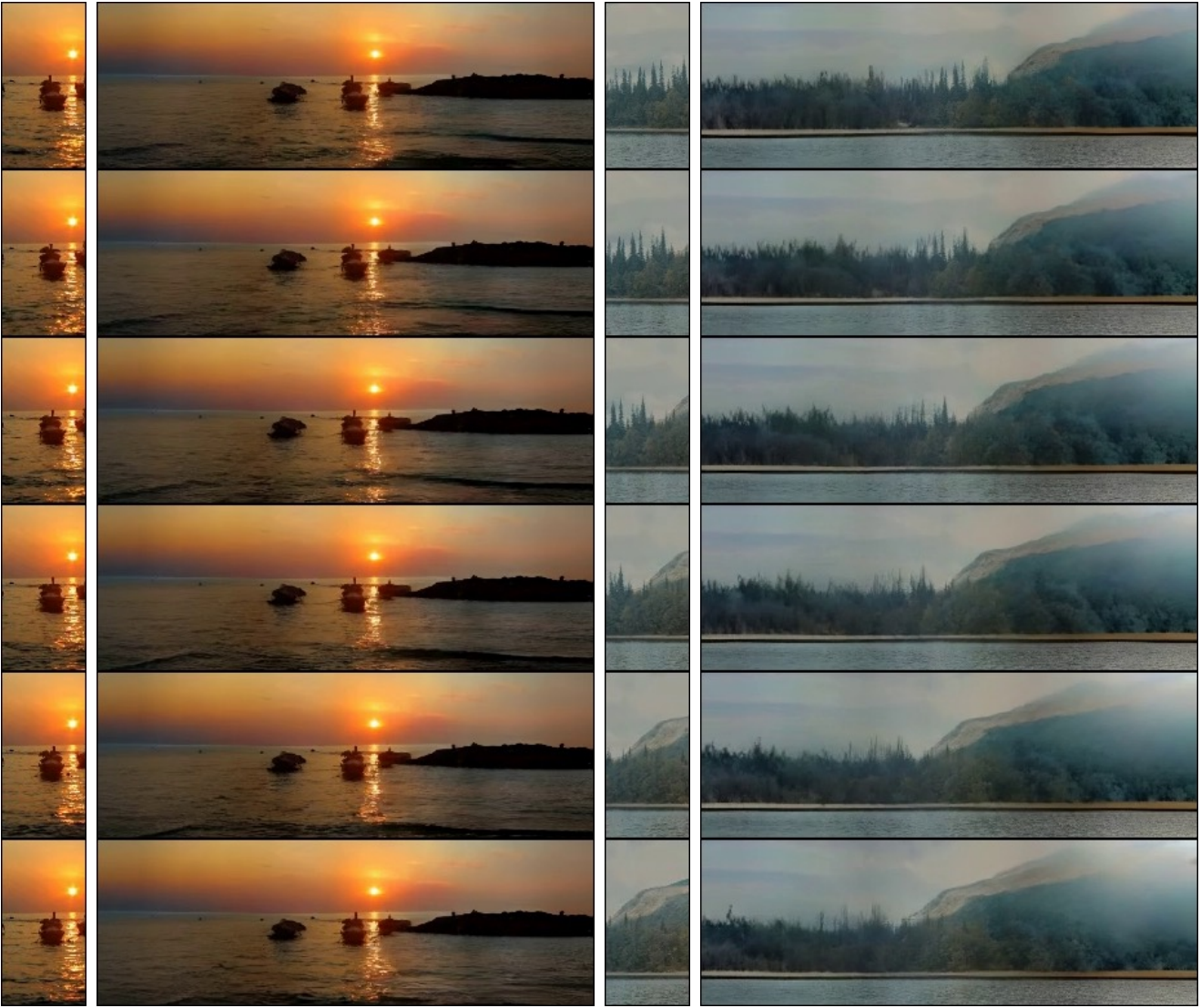}
\caption{\textbf{Ultrawide outpainting results}. Given a vertical slice of 64$\times$128, \modelname{} expands it into a panorama video of 384$\times$128 by doing vertical outpainting for 5 times on each side.}
\label{fig:wide_op2}
\end{figure*}

The diverse video generation tasks that \modelname{} is capable of can enable many useful applications.
For example, \cref{fig:wide_op1,fig:wide_op2} show a few untrawide outpainting samples by repeatedly performing the vertical outpainting task.
\modelname{} can easily generate nice large panorama videos given a small condition.



\clearpage
{\small
\bibliographystyle{ieee_fullname}
\bibliography{reference}
}